\documentclass[10pt,twocolumn,letterpaper]{article}

\usepackage{cvpr}
\usepackage{times}
\usepackage{epsfig}
\usepackage{graphicx}
\usepackage{amsmath}
\usepackage{amssymb}

\usepackage[pagebackref=true,breaklinks=true,letterpaper=true,colorlinks,bookmarks=false]{hyperref}

\cvprfinalcopy 

\def\cvprPaperID{2103} 

\ifcvprfinal\fi

\usepackage{graphicx}
\usepackage{xspace}
\usepackage{rotating}
\usepackage{multirow} 
\usepackage{cite}
\usepackage{url}
\usepackage{verbatim}

\usepackage{amsmath}
\usepackage{amssymb}
\usepackage{bbm} 
\usepackage{bm} 

\usepackage{pstricks}
\usepackage{pst-plot}
\usepackage[pstadd]{pst-sigsys}
\usepackage{pstricks-add}
\usepackage{multido}

\usepackage[marginclue,inline,final]{fixme}

\newcommand{\norm}[1]{\ensuremath{\lVert#1\rVert}}

\newcommand{\lspan}[1]{\ensuremath{\mathrm{span}}}
\newcommand{\by}[2]{\ensuremath{#1 \! \times \! #2}}

\newcommand{\xs}{\ensuremath{x}}        %

\newcommand{\vv}{\ensuremath{\mathbf{v}}}        %
\newcommand{\wv}{\ensuremath{\mathbf{w}}}        %
\newcommand{\xv}{\ensuremath{\mathbf{\xs}}}        %
\newcommand{\fig}{Fig.}

\makeatletter 
\DeclareRobustCommand\onedot{\futurelet\@let@token\@onedot}
\def\@onedot{\ifx\@let@token.\else.\null\fi\xspace}

\def\etal{et al\onedot} 
\makeatother

\usepackage{hyperref}

\begin{document}
\title{Untangling Local and Global Deformations in Deep Convolutional Networks
  for Image Classification and Sliding Window Detection}
\author{%
  George Papandreou\\
  Toyota Technological Institute at Chicago\\
  {\tt\small gpapan@ttic.edu}
  \and
  Iasonas Kokkinos and Pierre-Andr\'e Savalle\\
  Ecole Centrale Paris and INRIA\\
  {\tt\small [iasonas.kokkinos,pierre-andre.savalle]@ecp.fr}
}
	
\sloppy

\maketitle
\newcommand{\mycomment}[1]{}

\begin{abstract}
  Deep Convolutional Neural Networks (DCNNs) commonly use generic
  `max-pooling' (MP) layers to extract deformation-invariant features, but we
  argue in favor of a more refined treatment. First, we introduce {\em
    epitomic convolution} as a building block alternative to the common
  convolution-MP cascade of DCNNs; while having identical complexity to MP,
  Epitomic Convolution allows for parameter sharing across different filters,
  resulting in faster convergence and better generalization. Second, we
  introduce a Multiple Instance Learning approach to explicitly accommodate
  global translation and scaling when training a DCNN exclusively with class
  labels. For this we rely on a {\em `patchwork'} data structure that
  efficiently lays out all image scales and positions as candidates to a
  DCNN. Factoring global and local deformations allows a DCNN to `focus its
  resources' on the treatment of non-rigid deformations and yields a
  substantial classification accuracy improvement. Third, further pursuing
  this idea, we develop an efficient DCNN sliding window object detector that
  employs explicit search over position, scale, and aspect ratio. We
  provide competitive image classification and localization results on the
  ImageNet dataset and object detection results on the Pascal VOC 2007
  benchmark.
\end{abstract}

\section{Introduction}
\label{sec:intro}

Deep learning offers a powerful framework for learning increasingly complex
representations for visual recognition tasks. The work of Krizhevsky \etal
\cite{KSH13} convincingly demonstrated that deep neural networks can be very
effective in classifying images in the challenging Imagenet benchmark
\cite{DDSL+09}, significantly outperforming computer vision systems built on
top of engineered features like SIFT \cite{Lowe04}. Their success spurred a
lot of interest in the machine learning and computer vision
communities. Subsequent work has improved our understanding and has refined
certain aspects of this class of models \cite{ZeFe13b}. A number of different
studies have further shown that the features learned by deep neural networks
are generic and can be successfully employed in a black-box fashion in other
datasets or tasks such as image detection \cite{ZeFe13b, OuWa13, SEZM+14,
  GDDM14, RASC14, CSVZ14}.

The deep learning models that so far have proven most successful in image
recognition tasks, e.g.
\cite{SzegedyLJSRAEVR14,nin,SiZi14},
are feed-forward convolutional neural networks trained in a
supervised fashion to minimize a regularized training set classification error
by back-propagation. Their recent success is partly due to the availability of
large annotated datasets and fast GPU computing, and partly due to some
important methodological developments such as dropout regularization and
rectifier linear activations \cite{KSH13}. However, the key building blocks of
deep neural networks for images have been around for many years \cite{LBBH98}:
(1) Deep Convolutional Neural Networks (DCNNs) with small receptive fields that
spatially share parameters within each layer. (2) Gradual abstraction and
spatial resolution reduction after each convolutional layer as we ascend the
network hierarchy, typically via max-pooling \cite{RiPo99, JKRL09}

\begin{figure}[!tbp]
  \centering
  \begin{tabular}{c}
    \includegraphics[width=0.9\columnwidth]{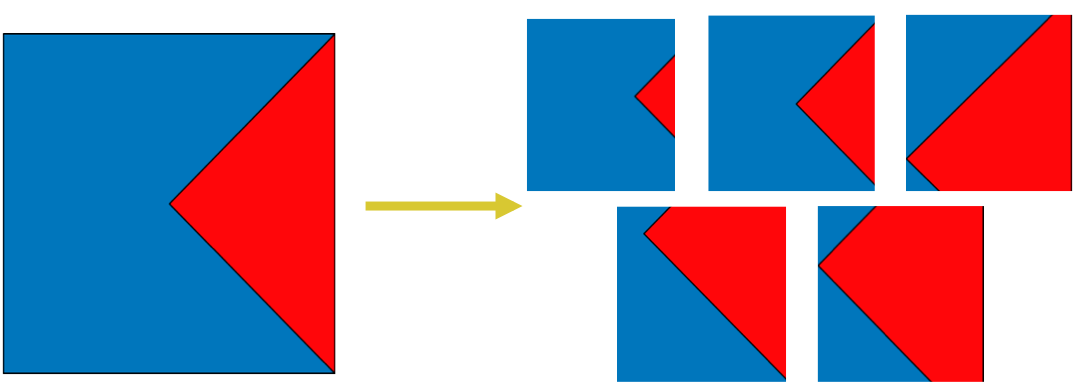}
  \end{tabular}
  \caption{Epitomes represent {\em a set} of small images through a common data
    structure: The epitome is a single, larger image which can produce
    several smaller filters by picking a position and cropping a small window.  While
    epitomes where originally employed for generative image modelling, in this
    work we propose to use  epitomes as an efficient method for parameter
    sharing in Deep Learning: Epitomes are used in hierarchical networks,
    and their parameters are learned discriminatively through
    back-propagation.}
  \label{fig:epitome_bag}
\end{figure}

This paper examines several aspects of invariance to deformations in the context of DCNNs for visual recognition. We treat both locally defined (non-rigid) and global 
deformations, demonstrating that a more refined treatment yields improved classification performance.

 Our first main contribution is
to build a deep neural network around the epitomic representation
\cite{JFK03}. The image epitome, illustrated in Figure~\ref{fig:epitome_bag},
is a data structure appropriate for learning translation-aware image
representations, naturally disentagling appearance and position modeling of
visual patterns. While originally developped for generative image modelling, we show here that the epitome data structure can be used to  train DCCNs discrimatively.

\begin{figure}[!tbp]
  \centering
  \begin{tabular}{c c}
    \includegraphics[width=0.45\columnwidth]{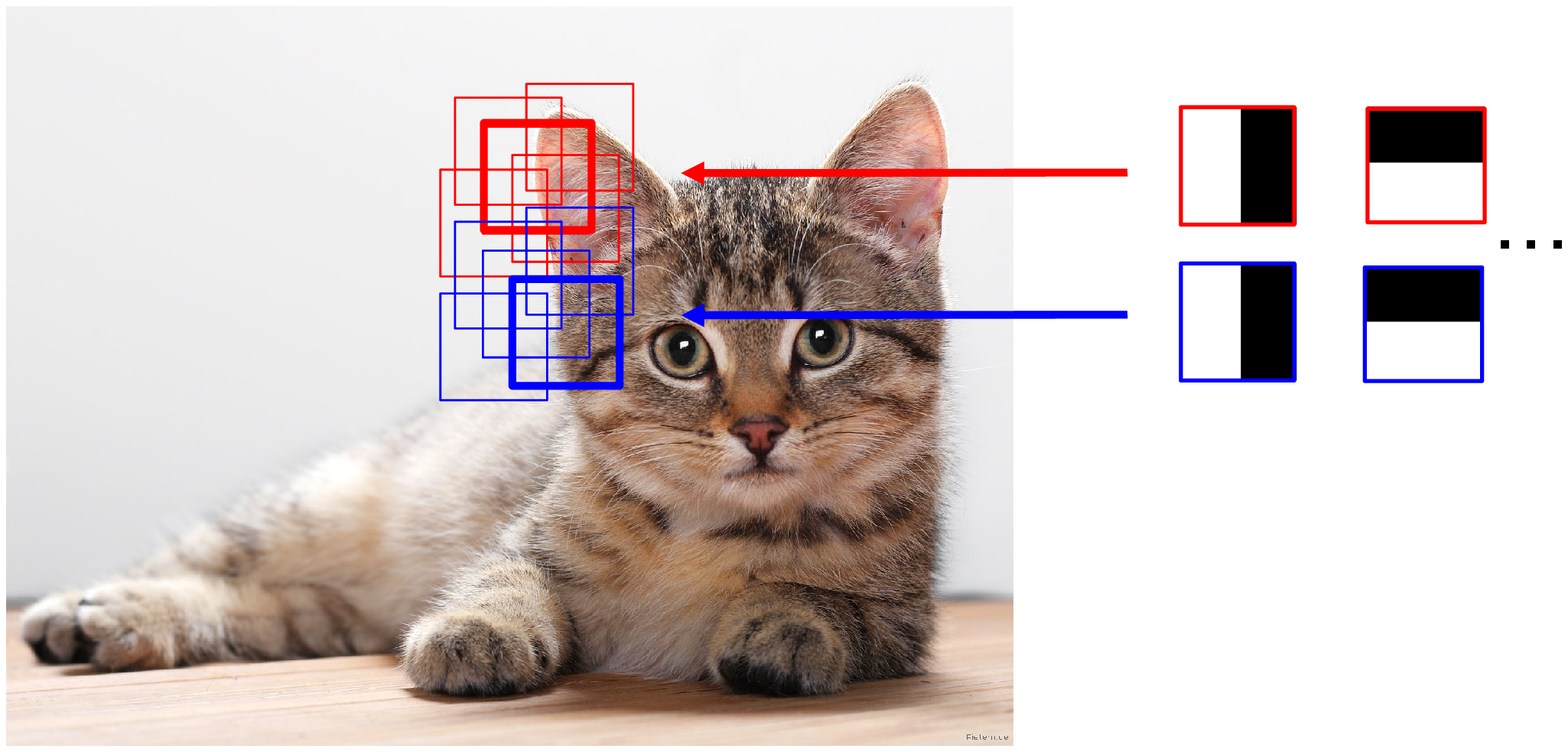}&
    \includegraphics[width=0.45\columnwidth]{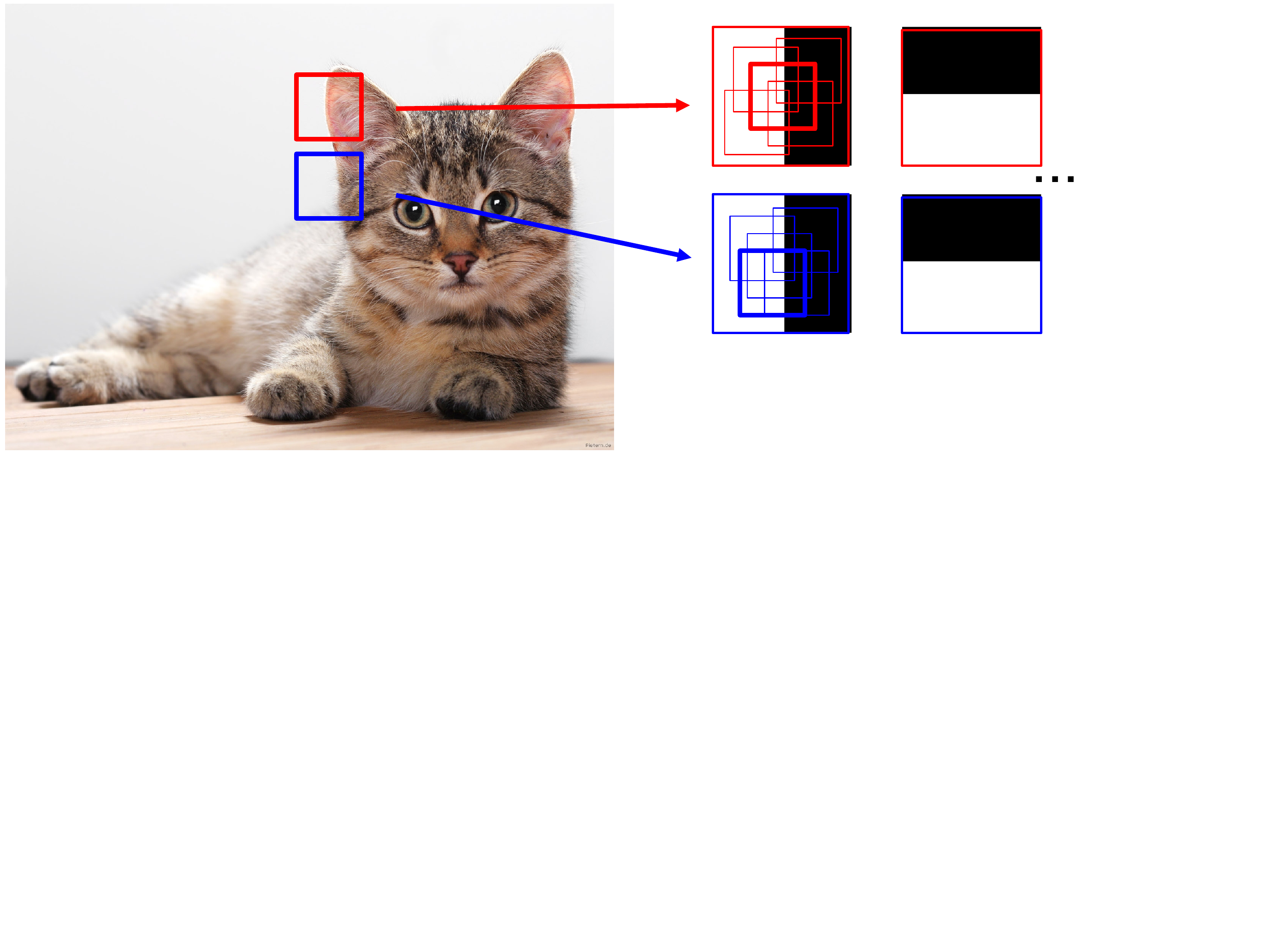}\\
    (a)&(b)
  \end{tabular}
  \caption{(a) Standard max-pooled convolution: For each filter we look for
    its best match within a small window in the data layer. (b) Proposed
    epitomic convolution (mini-epitome variant): For input data patches
    sparsely sampled on a regular grid we look for their best match in each
    mini-epitome.}
  \label{fig:epitome_diagram}
\end{figure}

For this, we substitute every pair of  convolution and max-pooling layers
typically used in DCNNs with an `Epitomic Convolution' layer. As illustrated in 
\fig~\ref{fig:epitome_diagram},  Epitomic Convolution in an input-centered dual alternative to the filter-centered standard
max-pooling. Namely, rather than searching in the input image (or layer) for the strongest response of a filter, we search across the set of filters encapsuled in a epitome for the strongest response to an image patch.
\mycomment{
for each regularly-spaced input data patch in the lower layer we search across
filters in the epitomic dictionary for the strongest response. In max-pooling
on the other hand, for each filter in the dictionary we search within a window
in the lower input data layer for the strongest response.}
Our Epitomic Convolution is designed to have the same computational and model complexity as Max-Pooling, while similar filters to share parameters. 

We use a dictionary of `mini-epitomes' at each network layer: each
mini-epitome is only slightly larger than the corresponding input data patch,
just enough to accomodate for the desired extent of position invariance. For
each input data patch, the mini-epitome layer outputs a single value per
mini-epitome, which is the maximum response across all filters in the
mini-epitome. In \cite{Papa14} we discuss another deep epitomic network
variant built on top of large epitomes which learns topographically organized
features.

Our second main contribution is to explicitly deal with object scale and
position when applying DCNNs to image classification. We show
that this can be done efficiently using a patchwork data structure in a  principled, consistent manner during both training and testing. 
 While
fusing classification results extracted from multiple image windows is
standard practice, we show that when incorporating multiple windows during training in a
Multiple Instance Learning (MIL) framework we obtain substantially larger
gains.

Finally, we  explore to what extent we can push scale and position search
 towards developing efficient and effective sliding window object
detectors on top of DCNNs. We accelerate sliding window detection by introducing 
 a simple method to reduce the effective size and receptive field of a DCNN pre-trained on ImageNet and show that by performing an explicit search over position, scale, and aspect ratios we can obtain  results that are comparable to the current-state-of-the-art while being substantially simpler and easier to train, as well as more efficient.

We quantitatively evaluate the proposed models primarily in image
classification experiments on the Imagenet ILSVRC-2012 large-scale image
classification task. We train the model by error back-propagation to minimize
the classification loss, similarly to \cite{KSH13}. We have experimented with
deep mini-epitomic networks of different architectures (Class A, B, and C). We
have carried out the bulk of our comparisons between epitomes and max-pooling
using comparable mid-sized networks having 6 convolutional and 3 fully
connected layers whose structure closely follows that in \cite{SEZM+14}
(Class-A). In these evaluations the deep epitomic network achieves 13.6\%
top-5 error on the validation set, which is 0.6\% better than the
corresponding conventional max-pooled convolutional network whose error rate
is 14.2\%, with both networks having the same computational cost. Note that
the error rate of the original model in \cite{KSH13} is 18.2\%, using however
a smaller network. We have also more recently experimented with larger deep
mini-epitomic networks (Class-B), which have the same number of levels but
more neurons per layer than those in Class-A. The plain deep epitomic net in
Class-B achieves an error rate of 11.9\%. When accompanied with explicit scale
and position search (patchwork variant), its error drops further down to
10.0\%. Finally, following \cite{SiZi14}, we have most recently experimented
with a very deep epitomic network with 13 convolutional and 3 fully connected
layers (Class-C). This achieves an even lower error rate of 10.0\% (without
scale and position search). All these performance numbers refer to
classification with a single network. We also find that the proposed epitomic
model converges faster, especially when the filters in the dictionary are
mean- and contrast-normalized, which is related to \cite{ZeFe13b}. We have
found this normalization to also accelerate convergence of standard max-pooled
networks. 

In \cite{Papa14} we further report additional sets of experiments. First, we
show that a deep epitomic network trained on Imagenet can be effectively used
as black-box feature extractor for tasks such as Caltech-101 image
classification. Second, we report excellent image classification results on
the MNIST and CIFAR-10 benchmarks with smaller deep epitomic networks trained
from scratch on these small-image datasets.

\paragraph{Related work}

Our model builds on the epitomic image representation \cite{JFK03}, which was
initially geared towards image and video modeling tasks. Single-level
dictionaries of image epitomes learned in an unsupervised fashion for image
denoising have been explored in \cite{AhEl08, BMBP11}. Recently, single-level
mini-epitomes learned by a variant of K-means have been proposed as an
alternative to SIFT for image classification \cite{PCY14}. To our knowledge,
epitomes have not been studied before in conjunction with deep models or
learned to optimize a supervised objective.

The proposed epitomic model is closely related to maxout networks
\cite{GWMCB13}. Similarly to epitomic matching, the response of a maxout layer
is the maximum across filter responses. The critical difference is that the
epitomic layer is hard-wired to model position invariance, since filters
extracted from an epitome share values in their area of overlap. This
parameter sharing significantly reduces the number of free parameters that
need to be learned. Maxout is typically used in conjunction with max-pooling
\cite{GWMCB13}, while epitomes fully substitute for it. Moreover, maxout
requires random input perturbations with dropout during model training,
otherwise it is prone to creating inactive features. On the contrary, we
have found that learning deep epitomic networks does not require dropout in
the convolutional layers -- similarly to \cite{KSH13}, we only use dropout
regularization in the fully connected layers of our network.

Other variants of max pooling have been explored before. Stochastic pooling
\cite{ZeFe13a} has been proposed in conjunction with supervised
learning. Probabilistic pooling \cite{LGRN09} and deconvolutional networks
\cite{ZKTF10} have been proposed before in conjunction with unsupervised
learning, avoiding the theoretical and practical difficulties associated with
building probabilistic models on top of max-pooling. While we do not explore
it in this paper, we are also very interested in pursuing unsupervised
learning methods appropriate for the deep epitomic representation.

\section{Deep Epitomic Convolutional Networks}
\label{sec:epitomes}

\subsection{Mini-Epitomic deep networks}

\begin{figure*}[!tbp]
  \centering
  \begin{tabular}{c c}
    \includegraphics[scale=1.4]{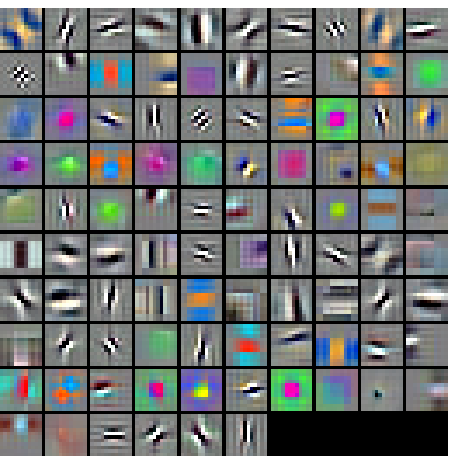}&
    \includegraphics[scale=1.4]{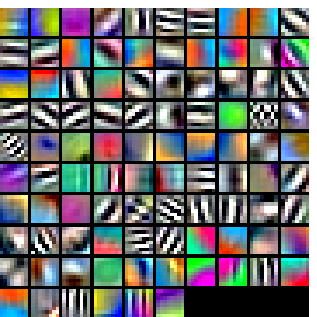}\\
    (a) & (b)
  \end{tabular}
  \caption{Filters at the first convolutional layer of: (a) Proposed
    \textsl{Epitomic} model with 96 mini-epitomes, each having size
    \by{12}{12} pixels. (b) Baseline \textsl{Max-Pool} model with 96 filters
    of size \by{8}{8} pixels each.}
  \label{fig:visualize_conv1}
\end{figure*}

We first describe a single layer of the mini-epitome variant of the proposed
model, with reference to \fig~\ref{fig:epitome_diagram}. In standard
max-pooled convolution, we have a dictionary of $K$ filters of spatial size
$\by{W}{W}$ pixels spanning $C$ channels, which we represent as real-valued
vectors $\{\wv_k\}_{k=1}^K$ with $W \cdot W \cdot C$ elements. We apply each
of them in a convolutional fashion to every $\by{W}{W}$ input patch
$\{\xv_i\}$ densely extracted from each position in the input layer which also
has $C$ channels. A reduced resolution output map is produced by computing the
maximum response within a small $\by{D}{D}$ window of displacements $p \in
\mathcal{N}_{input}$ around positions $i$ in the input map which are $D$
pixels apart from each other. 
The output map $\{z_{i,k}\}$ of standard max-pooled convolution has spatial
resolution reduced by a factor of $D$ across each dimension and will consist
of $K$ channels, one for each of the $K$ filters. Specifically:
\begin{equation}
  (z_{i,k},p_{i,k}) \leftarrow \max_{p \in \mathcal{N}_{image}}
  \xv_{i+p}^T \wv_k \,
  \label{eq:conv-max}
\end{equation}
where $p_{i,k}$ points to the input layer position where the maximum is
attained (argmax).

In the proposed epitomic convolution scheme we replace the filters with larger
mini-epitomes $\{\vv_k\}_{k=1}^K$ of spatial size $\by{V}{V}$ pixels, where $V
= W+D-1$. Each mini-epitome contains $D^2$ filters $\{\wv_{k,p}\}_{k=1}^K$ of
size $\by{W}{W}$, one for each of the $\by{D}{D}$ displacements $p \in
\mathcal{N}_{epit}$ in the epitome. We \emph{sparsely} extract patches
$\{\xv_i\}$ from the input layer on a regular grid with stride $D$ pixels. In
the proposed epitomic convolution model we reverse the role of filters and
input layer patches, computing the maximum response over epitomic positions
rather than input layer positions:
\begin{equation}
  (y_{i,k},p_{i,k}) \leftarrow \max_{p \in \mathcal{N}_{epitome}}
  \xv_i^T \wv_{k,p} \,
  \label{eq:epitomic-conv}
\end{equation}
where $p_{i,k}$ now points to the position in the epitome where the maximum is
attained. Since the input position is fixed, we can think of epitomic matching
as an input-centered dual alternative to the filter-centered standard max-pooling.

Computing the maximum response over filters rather than image
positions resembles the maxout scheme of \cite{GWMCB13}, yet in the proposed
model the filters within the epitome are constrained to share values in their
area of overlap.

Similarly to max-pooled convolution, the epitomic convolution output map
$\{y_{i,k}\}$ has $K$ channels and is subsampled by a factor of $D$ across
each spatial dimension. Epitomic convolution has the same computational cost
as max-pooled convolution. For each output map value, they both require
computing $D^2$ inner products followed by finding the maximum
response. Epitomic convolution requires $D^2$ times more work per input patch,
but this is exactly compensated by the fact that we extract input patches
sparsely with a stride of $D$ pixels.

Similarly to standard max-pooling, the main computational primitive is
multi-channel convolution with the set of filters in the epitomic dictionary,
which we implement as matrix-matrix multiplication and carry out on the GPU,
using the cuBLAS library.

It is noteworthy that conventional max-pooled convolution can be cast as
special case of epitomic convolution. More specifically, we can exactly
replicate max-pooled convolution with filters of size $\by{W}{W}$ and
$\by{D}{D}$ max-pooling using as epitomes zero-padded versions of the original
filters (with padding equal to $D-1$ pixels on each side of the filter) and
epitomic filter size equal to $W+D-1$. In that sense, epitomic convolution is
a generalization of max-pooling.

To build a deep epitomic model, we stack multiple epitomic convolution layers
on top of each other. The output of each layer passes through a rectified
linear activation unit $y_{i,k} \leftarrow \max(y_{i,k} + \beta_k, 0)$ and fed
as input to the subsequent layer, where $\beta_k$ is the bias. Similarly to
\cite{KSH13}, the final two layers of our network for Imagenet image
classification are fully connected and are regularized by dropout. We learn
the model parameters (epitomic weights and biases for each layer) in a
supervised fashion by error back propagation. We present full details of our
model architecture and training methodology in the experimental section.

\subsection{Optional mean and contrast normalization}
\label{sec:mean_con_norm}

Motivated by \cite{ZeFe13b}, we have explored the effect of filter mean
and contrast normalization on deep epitomic network training. More
specifically, we considered a variant of the model where the epitomic
convolution responses are computed as:
\begin{equation}
  (y_{i,k},p_{i,k}) \leftarrow \max_{p \in \mathcal{N}_{epitome}}
  \frac{\xv_i^T \bar{\wv}_{k,p}}{\norm{\bar{\wv}_{k,p}}_\lambda} \,
  \label{eq:epitomic-conv-norm}
\end{equation}
where $\bar{\wv}_{k,p}$ is a mean-normalized version of the filters and
$\norm{\bar{\wv}_{k,p}}_\lambda \triangleq (\bar{\wv}_{k,p}^T \bar{\wv}_{k,p}
+ \lambda)^{1/2}$ is their contrast, with $\lambda = 0.01$ a small positive
constant. This normalization requires only a slight modification of the
stochastic gradient descent update formula and incurs negligible computational
overhead. Note that the contrast normalization explored here is slightly
different than the one in \cite{ZeFe13b}, who only scale down the filters
whenever their contrast exceeds a pre-defined threshold.

We have found the mean and contrast normalization of
Eq.~\eqref{eq:epitomic-conv-norm} to significantly accelerate training not
only of epitomic but also max-pooled convolutional nets.

\subsection{Image Classification Experiments}

\begin{table*}[t]
\setlength{\tabcolsep}{3pt}
\begin{center}
\scalebox{1.00} {
\begin{tabular}{|l||c|c|c|c|c|c||c|c||c|}
  \hline  
  Layer           &   1 & 2 & 3 & 4 & 5 & 6 & 7 & 8   & Out \\
  \hline
  \hline
  Type            & conv +   & conv +   & conv & conv & conv & conv + & full + & full + & full \\
                  & lrn + max& lrn + max&      &      &      & max    & dropout& dropout&      \\ \hline
  Output channels &  96      & 192      & 256  & 384  & 512  & 512    & 4096   &  4096  & 1000 \\ \hline
  Filter size     &  8x8     & 6x6      & 3x3  & 3x3  & 3x3  & 3x3    & -    & -    & - \\ \hline
  Input stride    &  2x2     & 1x1      & 1x1  & 1x1  & 1x1  & 1x1    & -    & -    & - \\ \hline
  Pooling size    &  3x3     & 2x2      & -    & -    & -    & 3x3    & -    & -    & - \\ \hline
\end{tabular}
}
\caption{Architecture of the baseline \textsl{Max-Pool} convolutional network (Class-A).}
\label{tab:max_pool_net}
\end{center}
\end{table*}

\paragraph{Image classification tasks}

We have performed most of our experimental investigation with epitomes on the
Imagenet ILSVRC-2012 dataset \cite{DDSL+09}, focusing on the task of image
classification. This dataset contains more than 1.2 million training images,
50,000 validation images, and 100,000 test images. Each image is assigned to
one out of 1,000 possible object categories. Performance is evaluated using
the top-5 classification error. Such large-scale image datasets have proven so
far essential to successfully train big deep neural networks with supervised
criteria.


\paragraph{Network architecture and training methodology}

For our Imagenet experiments, we compare the proposed deep epitomic networks
with deep max-pooled convolutional networks. For fair comparison, we use as
similar architectures as possible, involving in both cases six convolutional
layers, followed by two fully-connected layers and a 1000-way softmax
layer. We use rectified linear activation units throughout the
network. Similarly to \cite{KSH13}, we apply local response normalization
(LRN) to the output of the first two convolutional layers and dropout to the
output of the two fully-connected layers. This is the Class-A of the models
we considered.

The architecture of our baseline \textsl{Max-Pool} network is specified on
Table~\ref{tab:max_pool_net}. It employs max-pooling in the convolutional
layers 1, 2, and 6. To accelerate computation, it uses an image stride equal
to 2 pixels in the first layer. It has a similar structure with the Overfeat
model \cite{SEZM+14}, yet significantly fewer neurons in the convolutional
layers 2 to 6. Another difference with \cite{SEZM+14} is the use of LRN, which
to our experience facilitates training.

The architecture of the proposed \textsl{Epitomic} network is specified on
Table~\ref{tab:mini_epitome_net}. It has exactly the same number of neurons at
each layer as the \textsl{Max-Pool} model but it uses mini-epitomes in place
of convolution + max pooling at layers 1, 2, and 6. It uses the same filter
sizes with the \textsl{Max-Pool} model and the mini-epitome sizes have been
selected so as to allow the same extent of translation invariance as the
corresponding layers in the baseline model. We use input image stride equal to
4 pixels and further perform epitomic search with stride equal to 2 pixels in
the first layer to also accelerate computation.

\begin{table*}[t]
\setlength{\tabcolsep}{3pt}
\begin{center}
\scalebox{1.00} {
\begin{tabular}{|l||c|c|c|c|c|c||c|c||c|}
  \hline  
  Layer           &   1 & 2 & 3 & 4 & 5 & 6 & 7 & 8   & Out \\
  \hline
  \hline
  Type            & epit-conv& epit-conv& conv & conv & conv & epit-conv & full + & full + & full \\
                  & + lrn    & + lrn    &      &      &      &           & dropout& dropout&      \\ \hline
  Output channels &  96      & 192      & 256  & 384  & 512  & 512       & 4096   &  4096  & 1000 \\ \hline
  Epitome size    &  12x12   & 8x8      &  -   &  -   &  -   & 5x5       & -    & -    & - \\ \hline
  Filter size     &  8x8     & 6x6      & 3x3  & 3x3  & 3x3  & 3x3       & -    & -    & - \\ \hline
  Input stride    &  4x4     & 3x3      & 1x1  & 1x1  & 1x1  & 3x3       & -    & -    & - \\ \hline
  Epitome stride  &  2x2     & 1x1      &  -   &  -   &  -   & 1x1       & -    & -    & - \\ \hline
\end{tabular}
}
\caption{Architecture of the proposed \textsl{Epitomic} convolutional network (Class-A).}
\label{tab:mini_epitome_net}
\end{center}
\end{table*}

We have also tried variants of the two models above where we activate the
mean and contrast normalization scheme of Section~\ref{sec:mean_con_norm} in
layers 1, 2, and 6 of the network.

We followed the methodology of \cite{KSH13} in training our models. We used
stochastic gradient ascent with learning rate initialized to 0.01 and
decreased by a factor of 10 each time the validation error stopped
improving. We used momentum equal to 0.9 and mini-batches of 128 images. The
weight decay factor was equal to $\by{5}{10^{-4}}$. Importantly, weight decay
needs to be turned off for the layers that use mean and contrast
normalization. Training each of the three models takes two weeks using a
single NVIDIA Titan GPU. Similarly to \cite{CSVZ14}, we resized the training
images to have small dimension equal to 256 pixels while preserving their
aspect ratio and not cropping their large dimension. We also subtracted for
each image pixel the global mean RGB color values computed over the whole
Imagenet training set. During training, we presented the networks with
$\by{220}{220}$ crops randomly sampled from the resized image area, flipped
left-to-right with probability 0.5, also injecting global color noise exactly
as in \cite{KSH13}. During evaluation, we presented the networks with 10
regularly sampled image crops (center + 4 corners, as well as their
left-to-right flipped versions).

\paragraph{Weight visualization}

We visualize in Figure~\ref{fig:visualize_conv1} the layer weights at the first
layer of the networks above. The networks learn receptive fields sensitive to
edge, blob, texture, and color patterns.

\paragraph{Classification results}

We report at Table~\ref{tab:imagenet_results} our results on the Imagenet
ILSVRC-2012 benchmark, also including results previously reported in the
literature \cite{KSH13, ZeFe13b, SEZM+14}. These all refer to the top-5 error
on the validation set and are obtained with a single network. Our best result
at 13.6\% with the proposed \textsl{Epitomic-Norm} network is 0.6\% better
than the baseline \textsl{Max-Pool} result at 14.2\% error. 
The improved performance that we got with the \textsl{Max-Pool} baseline
network compared to Overfeat \cite{SEZM+14} is most likely due to our use of
LRN and aspect ratio preserving image resizing. 


\begin{table*}[t]
\setlength{\tabcolsep}{3pt}
\begin{center}
\scalebox{0.9} {
\begin{tabular}{|l||c|c|c||c|c|c|c||c||c|}
  \hline
              & \multicolumn{3}{|c||}{Previous literature}  & \multicolumn{4}{|c||}{Class-A}             &  Class-B & Class-C \\
  \hline
  Model       & Krizhevsky  & Zeiler-Fergus  & Overfeat     & Max-Pool &  Max-Pool & Epitomic  & Epitomic & Epitomic & Epitomic\\
              & \cite{KSH13}& \cite{ZeFe13b} &\cite{SEZM+14}&          &  + norm   &           & + norm   &  +norm   & +norm   \\
  \hline
  Top-5 Error &   18.2\%    &    16.0\%      &  14.7\%      &  14.2\%  &  14.4\%   &\bf{13.7\%}&\bf{13.6\%}& 11.9\%  & 10.0\%  \\
  \hline
\end{tabular}
}
\caption{Imagenet ILSVRC-2012 top-5 error on validation set. All performance
  figures are obtained with a single network, averaging classification
  probabilities over 10 image crops (center + 4 corners, as well as their
  left-to-right flipped versions). Classes B and C refer to respectively
  larger and deeper models.}
\label{tab:imagenet_results}
\end{center}
\end{table*}

Further experimental results with the epitomic model on the Caltech-101 task
and on the MNIST and CIDAR-10 datasets are described in \cite{Papa14}. 



\paragraph{Mean-contrast normalization and convergence speed}

\begin{figure}[!tbp]
  \centering
  \includegraphics[width=0.8\columnwidth]{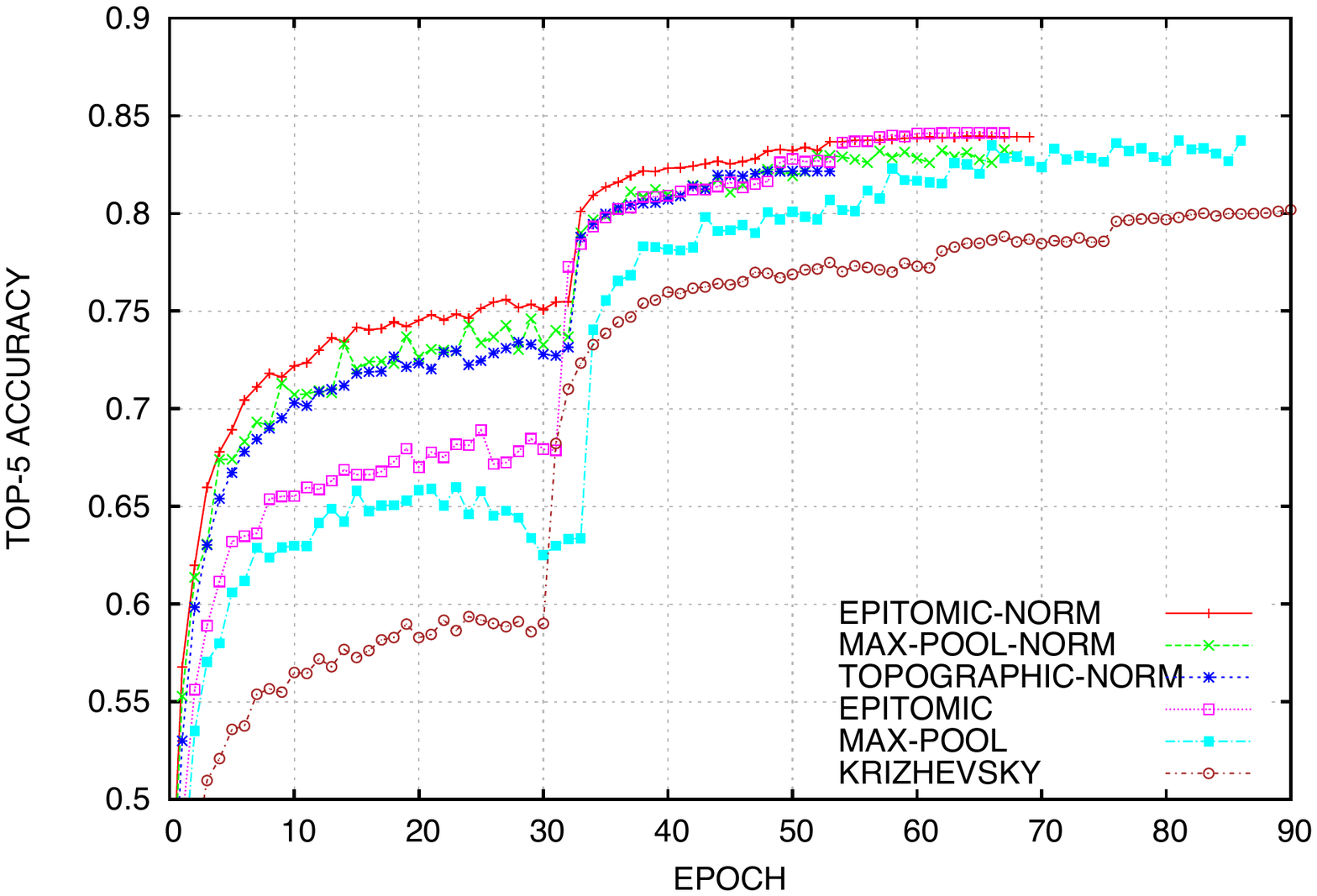}
  \caption{Top-5 validation set accuracy (center non-flipped crop only) for
    different models and normalization.}
  \label{fig:imagenet_optim}
\end{figure}

We comment on the learning speed and convergence properties of the different
models we experimented with on Imagenet. We show in
Figure~\ref{fig:imagenet_optim} how the top-5 validation error improves as
learning progresses for the different models we tested, with or without
mean+contrast normalization. For reference, we also include a corresponding
plot we re-produced for the original model of Krizhevsky \etal
\cite{KSH13}. We observe that mean+contrast normalization significantly
accelerates convergence of both epitomic and max-pooled models, without
however significantly influencing the final model quality. The epitomic models
converge faster and are stabler during learning compared to the max-pooled
baselines, whose performance fluctuates more.

\paragraph{Experiments with larger and deeper epitomic networks}

We have also experimented with larger (Class-B) and very deep (Class-C)
versions of the proposed deep epitomic networks. The large Class-B network has
the same number of levels but more neurons per layer than the networks in
Class-A. It achieves an error rate of 11.9\%. 

Inspired by the success of the top-performing methods in this year's Imagenet
competition, we have also very recently experimented with a very deep network
having 13 convolutional and 3 fully connected layers, which roughly follows
the architecture of the 16 layer net in \cite{SiZi14}. Our Class-C deep
epitomic network achieves 10.0\% error rate in the Imagenet task. The
state-of-art 16 layer net in \cite{SiZi14} (without multi-scale
training/testing) achieves an even lower 9.0\% error rate, but using a more
sophisticated procedure for aggregating the results of multiple image crops (in
place of our simple 10-view testing procedure). As extra evidence to the
improved robustness of training our deep epitomic networks, it is noteworthy
to mention that we managed to train our very deep epitomic net starting from a
random initialization, while \cite{SiZi14} had to bootstrap their very deep
networks from shallower ones.

\section{Explicit Scale and Position Search in DCNNs}

The effects of object translation and scaling impede the training of accurate classifiers from datasets that lack bounding box annotations - as is the case in the ImageNet classification Challenge. The Epitomic Convolution/Max-Pooling modules allow us to extract an image representation that is invariant to signal deformations taking place at the pooling region's scale; when used in alternation with feature downsampling (`striding') this can result in invariance to increasingly large-scale signal transformations, eventually dealing with global position and scale changes.


\newcommand{\sample}{X}
\newcommand{\Sample}{\mathbf{X}}
\newcommand{\cllabel}{y}

\newcommand{\refeq}[1]{Eq.~\ref{#1}}

\newcommand{\ba}{\begin{eqnarray}}
	\newcommand{\ea}{\end{eqnarray}}

A better treatment of deformations can be achieved by factoring deformations
into local (non-rigid) and global (translation/scale) changes. We can then
simulate the effect of the latter during training and testing, by transforming the input images, obtaining us an additional hold on the problem. We now describe how this idea can be exploited in the setting of Multiple Instance Learning (MIL).

\subsection{MIL-based training of DCNNs}
Considering a binary classification problem with $N$ image-label pairs
$\mathcal{S} = \{(\sample_i,y_i)\}, i=1,\ldots,N$, training aims at minimizing the following criterion:
\ba
C(f,\mathcal{S}) = \sum_{i=1}^{N} l(y_i,f(\sample_i)) + R(f), \label{eq:original} 
\ea
where $f$ is the classifier, $l(y,f(\sample))$ is the  loss function and $R$ is a regularizer. 

{\em Dataset augmentation} amounts to turning an image  $\sample_i$ into a set of images $\Sample_i = \{\sample^1_i,\ldots,\sample^K_i\}$  by transforming $\Sample_i$ synthetically; e.g. considering $T$  translations and $S$ scalings yields a set with $K = T S$ elements. 
The most common approach to using Dataset augmentation consists in   treating each element of $\Sample_i$  as a new training sample, i.e.  substituting the loss $l(y,f(\sample_i))$ in \label{eq:original} by the sum of the classifier's loss on all images:
\ba
L(y_i,\Sample_i) \doteq \sum_{k=1}^K l(y_i,f(\sample^k_i)).
\label{eq:sum1}
\ea
This corresponds to the dataset augmentation technique use  e.g. in \cite{Howa13}.
A recently introduced alternative is the `sum-pooling' technique used in \cite{SiZi14}, which can be understood as using the following loss:
\ba
L(y_i,\Sample_i) =  l(y_i,\frac{1}{K}\sum_{k=1}^K f(\sample^k_i)), \label{eq:sum2}
\ea
which averages the classifier's score over translated versions of the input image. 
The summation used in both of these approaches favors classifiers  that consistently score highly on positive samples, irrespective of the object's position and scale - this is when the loss is minimized. As such, these classifiers  can be seen as pursuing the invariance of $f$. 

There is however a  tradeoff between invariance and classification accuracy \cite{Varma}. Even though pursuing invariance accounts for the effects of transformations, it does not make the classification task any easier: the training objective aims at a classifier that would allow all transformed  images to make it through its `sieve'. Understandably, a classifier that only considers centered objects of a fixed scale can acchieve higher accuracy: this would allow us to devotes all modelling resources to the treatment of local deformations. 

For this, we  let our classifier  `choose' its preferred transformation 
In particular, we define the loss function to be:
\ba
L(y_i,\Sample_i) =  l(y_i,\max_{k} f(\sample^k_i)), \label{eq:mil}
\ea
which amounts to letting the classifier choose the transformation that maximizes its response on a per-sample basis, and then accordingly penalizing that response. In particular the loss function requires that the classifier's response is large on at least one  position for a positive sample - and small everywhere for a negative. 

This idea amounts to the simplest case of  Multiple Instance Learning \cite{diet97}: $\Sample_i$ can be seen as a {\em bag of features} and the individual elements of $\Sample_i$ can be seen as {\em instances}.  
For the particular case of the hinge loss function, this would lead us to the latent-SVM training objective \cite{FGMR10,AndrewsHT02}.

Using this loss function during training amounts to treating the object's
position and scale as a latent variable, and performing alternating
optimization over the classifier's score function. During testing we perform
a search over transformations and keep the best classifier score, which can be understood as maximizing
over the latent transformation variables.
The resulting score
$F(\Sample_i) = \max_{k} f(\sample^k_i)$
is transformation-invariant, but is built on top of a classifier tuned for a single scale- and translation- combination. The MIL setting allows us to 
train and test our classifiers consistently, using the same set of image
translations and scalings.

\begin{table}[t]
\begin{center}
\begin{tabular}{|l||c||c|}
  \hline  
  Model       & Epitomic   &    Epitomic   \\
              & (Class-B)  &  (patchwork)  \\ \hline
  Top-5 Error &    11.9\%  &     10.0\%    \\
  \hline
\end{tabular}
\caption{Imagenet ILSVRC-2012 top-5 error on validation set. We compare the
  Class-B mean and contrast normalized deep epitomic network of
  Table~\ref{tab:imagenet_results} with its Patchwork fine-tuned version that
  also includes scale and position search.}
\label{tab:imagenet_results2}
\end{center}
\end{table}

\subsection{Efficient Convolutional Implementation}
\newcommand{\pyr}{\mathcal{I}}
We now turn to practical aspects of integrating  MIL  into DCNN training. DCNNs are commonly trained with input images of a fixed size, $q\times q$. For an arbitrarily-sized input image  $I$, if  we 
denote by $\pyr(x,y,s)$ its image pyramid, naively computing the maximization in \refeq{eq:mil} would require to cropping many $q\times q$ boxes from $\pyr(x,y,s)$, evaluating $f$ on them, yielding $f(x,y,s)$, and then penalizing  $l(y,\max_{x,y,s} f(x,y,s))$ during training (we ignore downsampling and boundary effects for simplicity). Doing this would require a large amount of GPU memory, communication and computation time. Instead, by properly modifying the input and architecture of our network
we can  share computation to efficently implement exhaustive search during training and testing. 

\begin{figure}[!tbp]
	\centering
	\begin{tabular}{c}
		\includegraphics[width=\columnwidth]{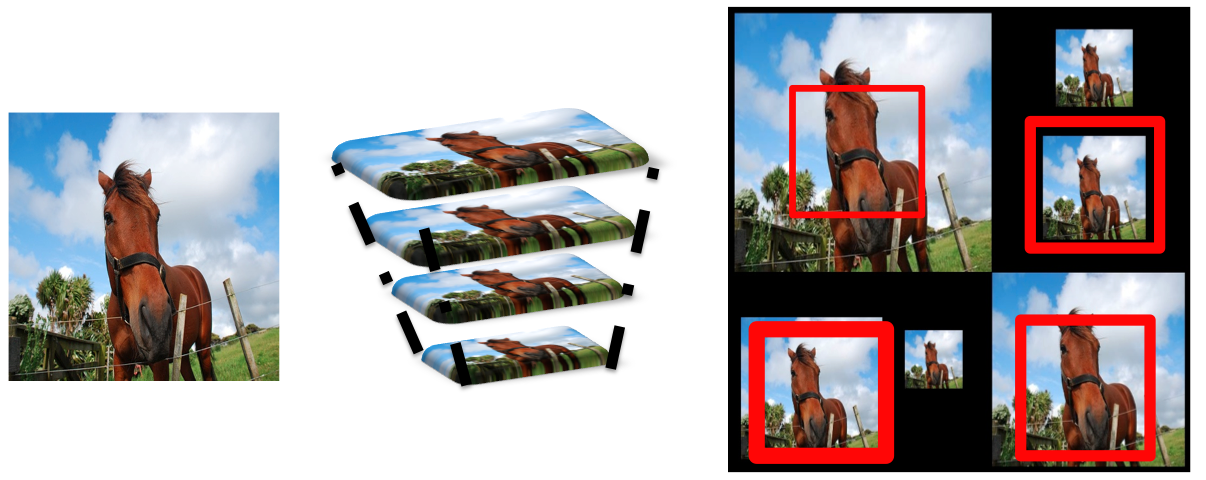}
	\end{tabular}
	\caption{We use the image patchwork technique to efficiently implement scale and position search in DCNN training: an image  pyramid is unfolded into a image `patchwork', where sliding a fixed-size window amounts to search over multiple positions and scales. The maximum of the classifier's score on all such  windows is efficiently gathered by max-pooling the DCNN's top-layer responses, and is used to accommodate scale and position changes during both training and testing.
		}
	\label{fig:patchwork}
\end{figure}


For this, we first  draw inspiration from the `image patchwork' technique
introduced in \cite{Dubout} and exploited in DCNNs by \cite{iandola}. The
technique consists in embedding a whole image pyramid $\pyr(x,y,s)$, into a
single, larger, patchwork image $P(x',y')$; any position $(x',y')$ in $P$
corresponds to a $(x,y,s)$ combination in $I$.  This was originally conceived
as a means of accelerating multi-scale FFT-based convolutions in \cite{Dubout}
and convolutional feature extraction in \cite{iandola}. Instead we view it a
stepping stone to implementing scale and position search during DCNN training.

In particular,  we treat the last, fully-connected, layers of a DCNN as
$1\times1$ convolution kernels; we can then obtain the $f(x,y,s)$ score
described  above by providing $P$ as input to our network, since the output of
our network's final layer at any position $(x',y')$  will correspond to the
output corresponding to a $q\times q$ square cropped around $(x,y,s)$. This
allows us to incorporate the $\max$ operation used in MIL's training
criterion, \refeq{eq:mil}, as an additional max-pooling layer situated on top
of the network's original score function. This makes it possible to
efficiently incorporate global scale and position search in a seamless manner
within DCNN training.

\subsection{Image Classification Results}

We have experimented with the scheme outlined above in combination with our
Deep Epitomic Network (Class-B variant) presented in the previous Section.  We
use a \by{720}{720} patchwork formed from 6 different image scales (square
boxes with size 400, 300, 220, 160, 120, and 90 pixels). We have resized all
train/test images to square size, changing their aspect ratio if needed. We
have initialized this scale/position search net with the parameters of our
standard Class-B epitomic net. We fine-tuned the network parameters for about
an epoch on the Imagenet train set, following the training methodology
outlined earlier.

We have obtained a substantial further decrease in the testing error rates,
cutting the top-5 error rate from 11.9\% down to 10.0\%, as shown in
Table~4. This reduction in error rate is competitive with the best reduction obtained by  more complicated techniques involving many views for evaluation \cite{SiZi14,SiZi14}, while also allowing for consistent end-to-end training and testing. 


The network outlined above also provides cues for the
scale and position of the dominant object in the image. A simple fixed mapping
of the ``argmax'' patchwork position in the last max-pooling layer (computed
by averaging the bounding box positions in the training set) yields 48.3\%
error rate in the Imagenet 2012 localization task without incurring any extra
computation.

\section{Sliding Window Object Detection with DCNNs}
\label{sec:detection}


The success of explicit position and scale search in image classification
suggests using DCNNs for sliding-window detection; even though for pedestrians
excellent results have been reported in \cite{yangW13} recent works on
combining convolutional networks \cite{SEZM+14}, or sliding window detectors
with CNN features \cite{iandola,savalle} still lag behind the current
state-of-the-art techniques \cite{GDDM14,HeZR014,DeepId}.
 
Starting from the RCNN work of \cite{GDDM14}, all such techniques compute DCNN
features `on-demand' for a set of 1000-2000 image regions delivered by
selective search \cite{ssearch}, and apply a separately trained SVM detection
on top. This approach has recently been shown to deliver compelling detection
results; most recently, \cite{GDDM14} have shown that combining RCNNs with the
network of \cite{SiZi14} pushes the mean AP performance on Pascal VOC 2007 to
$66\%$ ($62\%$ without bounding box regression), but at the cost of 60 seconds per image (acceleration techniques such
as Spatial Pyramid Pooling \cite{HeZR014} can still be applied, though).
 
Despite the performance gap of sliding window detection to RCNNs, we consider
sliding window detection simpler, and potentially more amenable to analysis
and improvement. 
 
\subsection{Explicit search over aspect ratios}

Our basic object detection system uses  as
input to a DCNN an image patchwork that includes 11
scales logarithmically sampled, from 2 times down to 1/6 of the image size.
Unlike recent works  \cite{iandola,savalle} that only
operate with the first five, `convolutional' layers of a deeper network,
 in our case we set the fully-connected layers operate convolutionally, and use
the DCNN class scores for proposing square bounding boxes as object detection
proposals. This processes a typical PASCAL VOC 2007 image in less than 1 sec
on a Titan Tesla K40. 

This square bounding box detector (without any bounding box regression
post-processing) yields a mean Average Precision of $43.0\%$ on Pascal VOC
2007. To further analyze which of this system's errors are due to constraining
the bounding box proposals to be square, we investigated the system's
performance in the presence of an `oracle' that provides the optimal aspect
ratio (using the ground truth annotations) for any given square bounding box
proposal. We found  this oracle bounding box prediction to increase
performance to $56.7\%$. This indicates that square box prediction is
insufficient for achieving competitive object detection results.

Once again, we pursue an explicit search approach to directly estimate the
aspect ratio of bounding box proposals without needing to resort to an
oracle. We account for aspect ratio variability by scaling the image along a
single dimension. For instance, scaling an image by $.5$ vertically means that
a vertically-elongated \by{100}{200} region appears as a \by{100}{100} square
region in the transformed image. Our square detector can then find this
preferable to other, differently scaled versions of the same region, and
thereby hint at the right object ratio. Aspect ratios that receive a lower
score are eliminated at the final nonmaximum suppression stage. We account for
aspect ratio during both testing and training (see below).

We perform this operation for 5 distinct aspect ratios, spanning the range of
$[1/3,3]$ with a geometric progression, as illustrated in
Figure~\ref{fig:pw}. This is applied at the whole patchwork level -- sliding
window detection on these patchworks then amounts to a joint search over
scale, position, and aspect ratio. This is more time-demanding (requiring 5
times more computation) but still quite fast, requiring about 5 secs on a
Tesla K40 for an average Pascal VOC 2007 image, and yields non-square bounding
box predictions without resorting to an oracle.

The related detection results are shown in Table~\ref{tab:voc2007}. Aspect
ratio search yields a very competitive mean Average Precision score of
$56.4\%$ (without any bounding box regression post-processing). This is only
slightly lower than the oracle-based score $56.7\%$, indicating that also
normalizing for aspect ratio during training leads to better object models.

It is noteworthy that our $56.4\%$ mAP result is significantly better than the
$46.9\%$ mAP result reported very recently by \cite{WEF14}. Our better results
should be attributed to our different training procedure (detailed below), our
explicit search over aspect ratios, and the use of a more powerful DCNN
classifier. Our system is also significantly simpler than \cite{WEF14} (which
also integrates deformable part models and includes non maximum suppression
during training). We anticipate that integrating these components into our
system could further increase performance at a higher computational cost.

We observe that our system's average performance ($56.4\%$ mAP) is  still below the one obtained by \cite{GDDM14} when using the network of \cite{SiZi14} ($62.2\%$ mAP without bounding box regression). This can be attributed to several aspects of our system, including (i) using a smaller number of network parameters (detailed below) (ii) performing the detection with smaller windows (detailed below)
(iii) not using a retraining stage with the hinge loss and  hard-negative mining, as \cite{GDDM14} do, and (iv)  missing out on regions found by Selective Search. We are confident that factor (iv) is the least important - having experimentally verified that the recall of bounding boxes is systematically better according to our pyramid's hypothesized positions, rather than the boxes delivered by Selective Search. We are currently investigating the effects of factors (i)-(iii).
Still, we consider that our system has an edge on the efficiency side: our detector requires approximately 5 seconds to consider all position, scale and aspect ratio combinations, while the  
 system of \cite{GDDM14} with the network of \cite{SiZi14} requires approximately 60 seconds, on identical hardware (a Tesla K40 GPU).
  
\begin{table*}[t]\scriptsize
\setlength{\tabcolsep}{3pt}
\begin{center}
\begin{tabular}{|c|c*{19}{|c}||c|}
\hline
VOC 2007 test & aero & bike & bird & boat & bottle  & bus & car & cat & chair & cow & table & dog & horse & mbike & person & plant & sheep & sofa & train & tv   & mAP\\
\hline\hline
Our work (VGG)& 64.4 & 72.1 & 54.6 & 40.4 & 46.5    & 66.2& 72.9& 58.2& 31.8  &69.8 & 31.8  & 59.3& 71.1  & 68.3  & 64.7   & 31.0  & 55.0  & 49.8 & 55.3  & 64.4 & 56.4\\
\hline
RCNN7 \cite{GDDM14} (VGG) & 71.6 &  73.5 & 58.1 & 42.2 & 39.4 & 70.7  & 76.0 &  74.5 & 38.7 & 71.0 & 56.9 & 74.5 & 67.9 &  69.6 & 59.3 & 35.7 & 62.1 &  64.0 & 66.5 & 71.2 &  62.2\\\hline
RCNN7 \cite{GDDM14} (UoT) &  64.2 & 69.7 & 50.0 & 41.9 & 32.0 & 62.6 & 71.0 & 60.7 & 32.7 & 58.5 & 46.5 & 56.1 & 60.6 & 66.8 & 54.2 & 31.5 & 52.8 & 48.9 & 57.9 & 64.7 & 54.2\\\hline
DPM \cite{WEF14} (UoT)& 49.3 & 69.5 & 31.9 & 28.7 & 40.4 & 61.5 & 61.5 & 41.5 & 25.5 & 44.5 & 47.8 & 32.0 & 67.5 &  61.8 & 46.7 & 25.9 & 40.5 & 46.0 & 57.1 & 58.2 & 46.9\\\hline
 \end{tabular}
 \caption{Detection average precision (\%) on the PASCAL VOC 2007 test set,
   using the proposed CNN sliding window detector that performs explicit
   position, scale, and aspect ratio search. We compare to the RCNN architecture of \cite{GDDM14}
   and the end-to-end trained DPMs of \cite{WEF14}. In parenthesis we indicate the DCNN used for detection: UoT is the University of Toronto DCNN \cite{KSH13} and VGG is the DCNN of Oxford's Visual Geometry Group \cite{SiZi14}.
   }
 \label{tab:voc2007}
 \end{center}
\end{table*}

\newcommand{\frmt}{}
\newcommand{\sz}{.62}
\newcommand{\szh}{.35}
\begin{figure*}
\centering
\begin{tabular}{ccc}
\includegraphics[width=\szh\columnwidth,height=\szh\columnwidth]{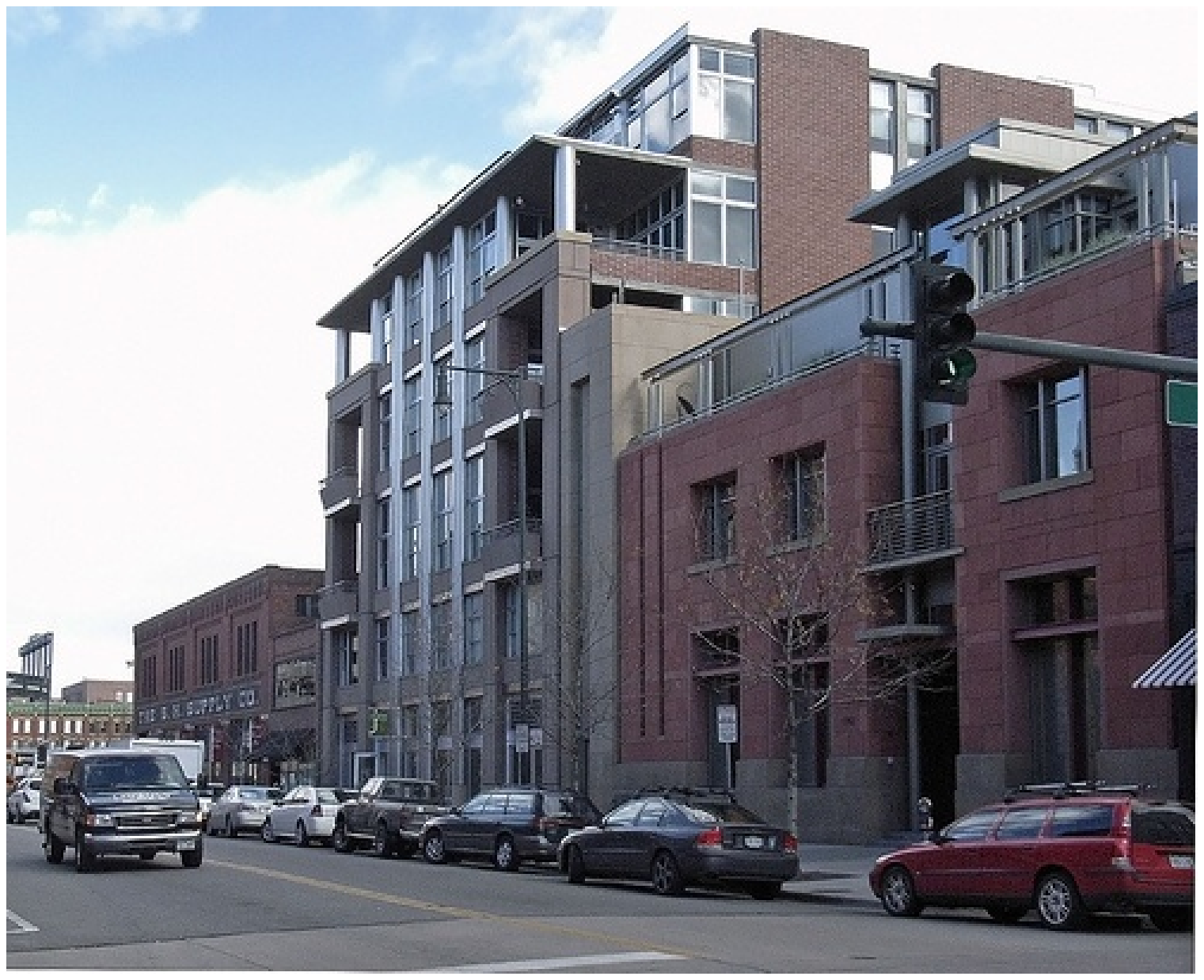}
&
\includegraphics[width=\sz\columnwidth,height=\sz\columnwidth]{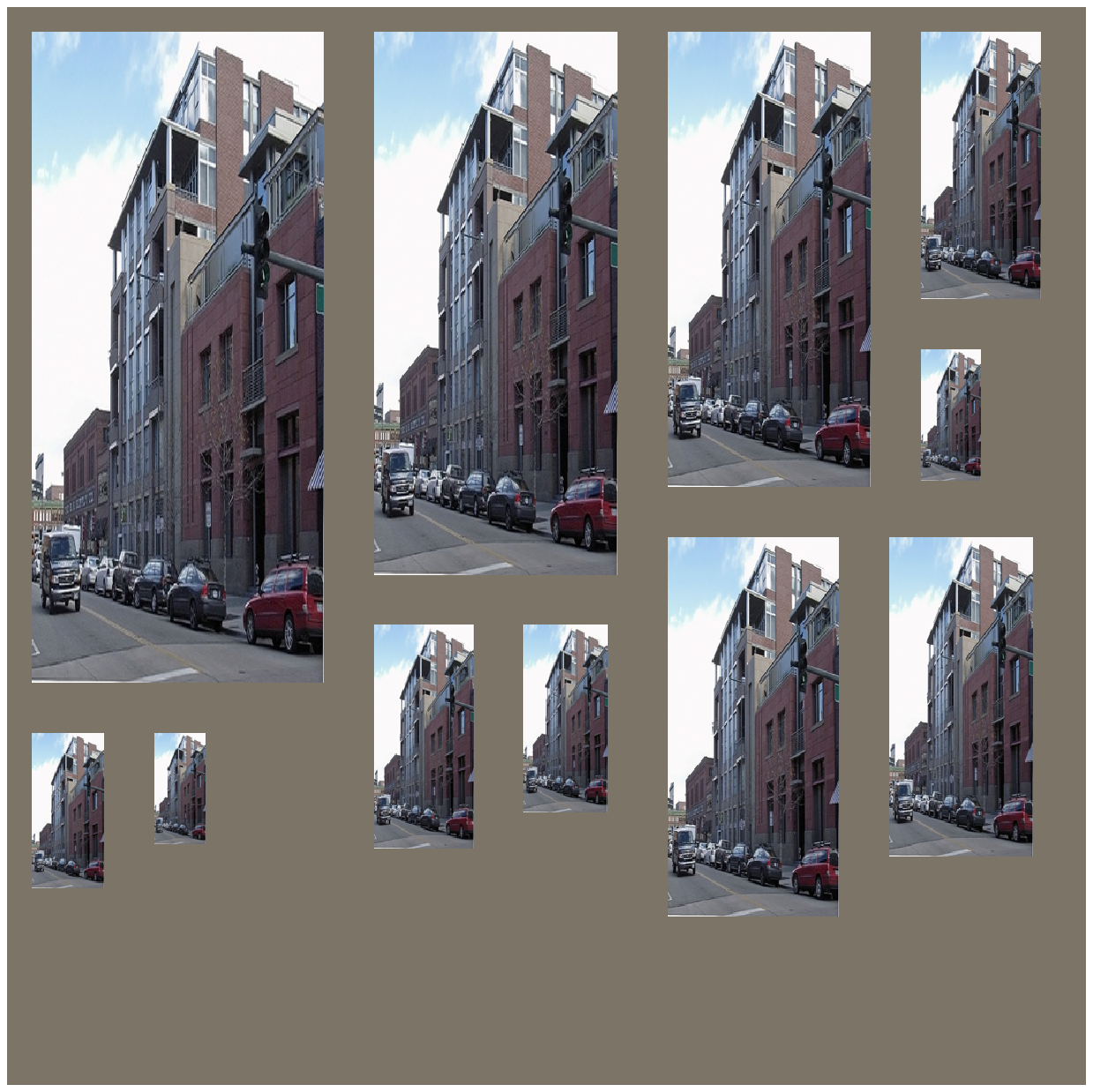}&
\includegraphics[width=\sz\columnwidth,height=\sz\columnwidth]{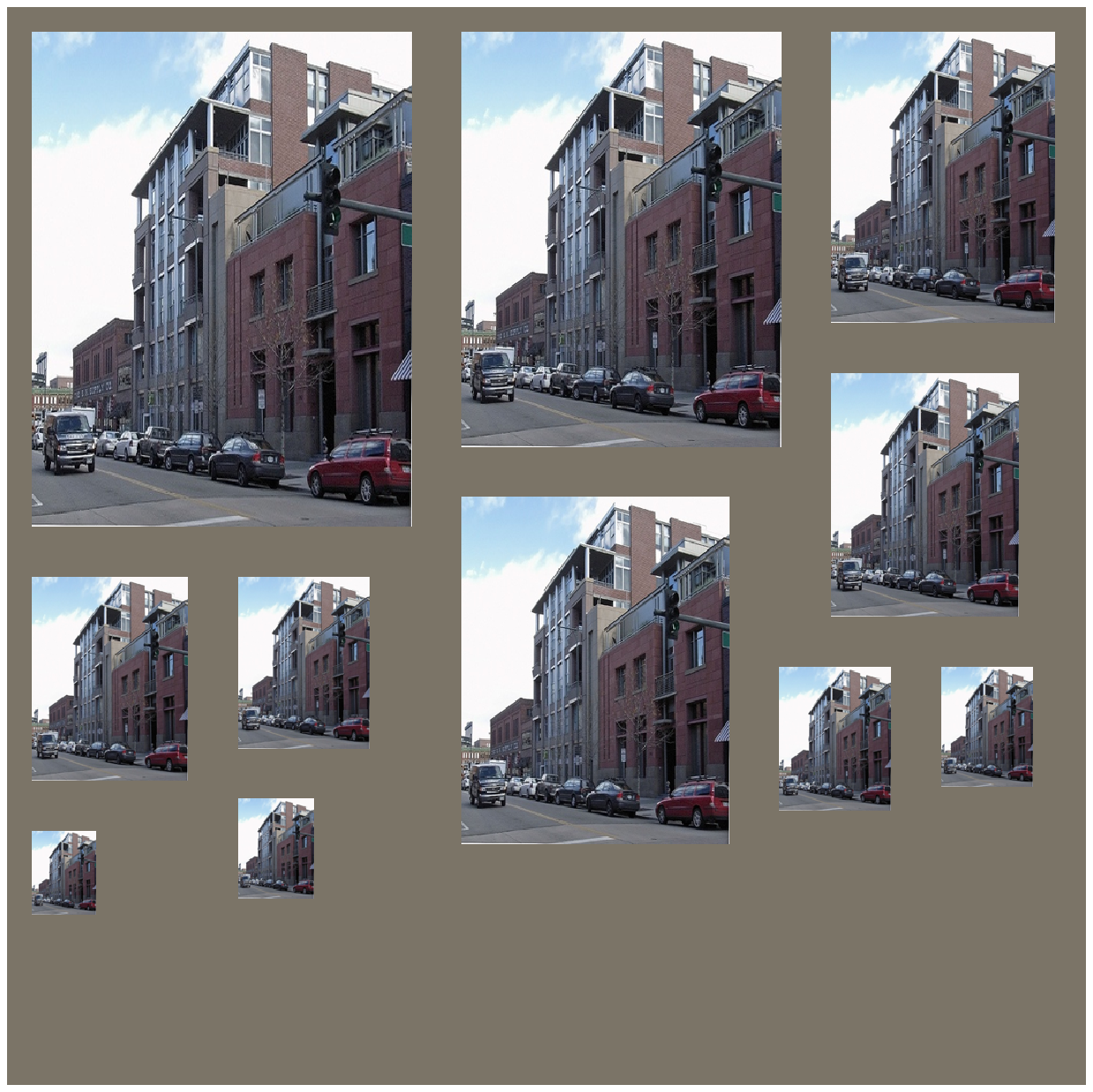}\\
Input image & Patchwork, $\alpha = 0.33$ & Patchwork, $\alpha = 0.57$ \\
&
\includegraphics[width=\sz\columnwidth,height=\sz\columnwidth]{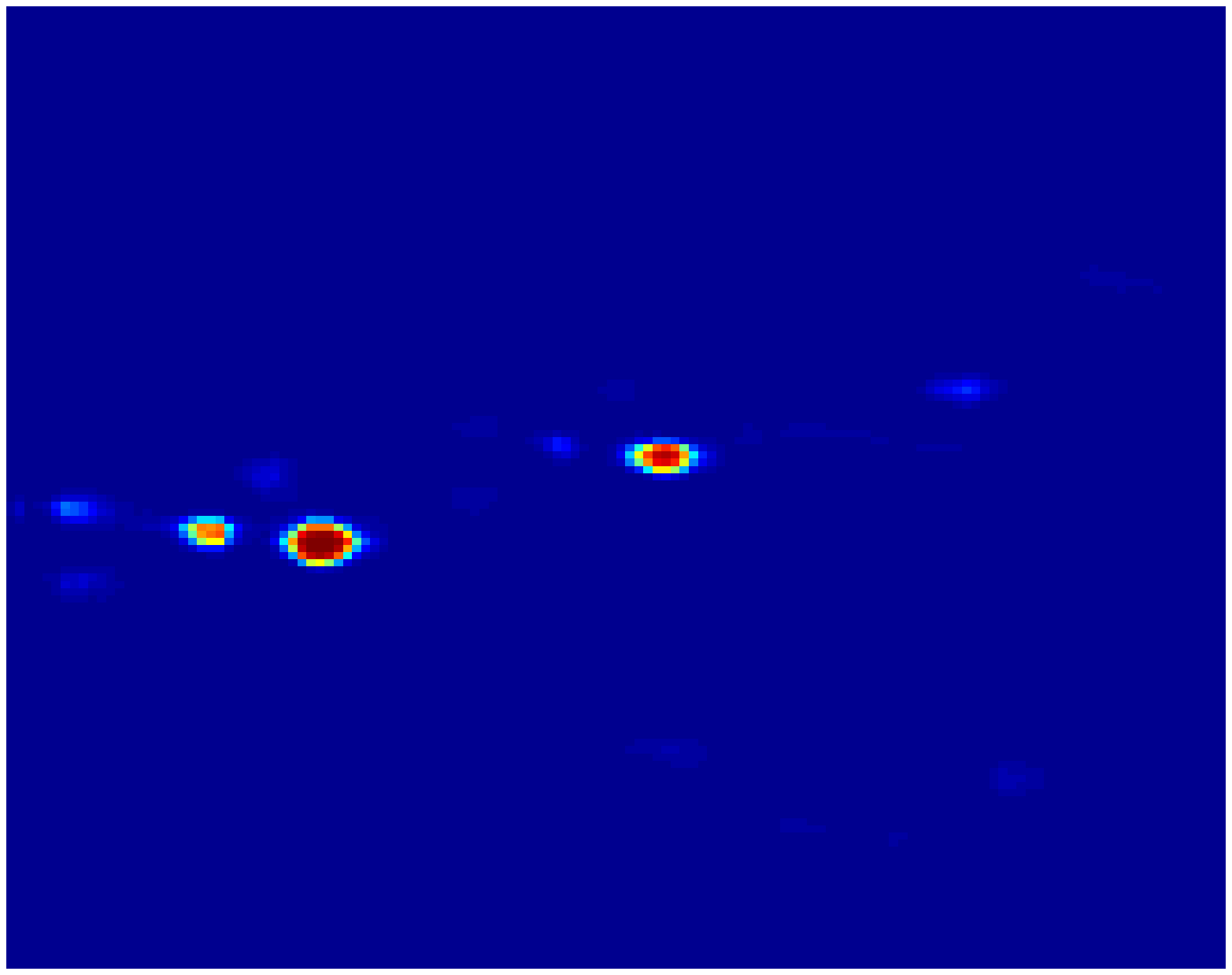}&
\includegraphics[width=\sz\columnwidth,height=\sz\columnwidth]{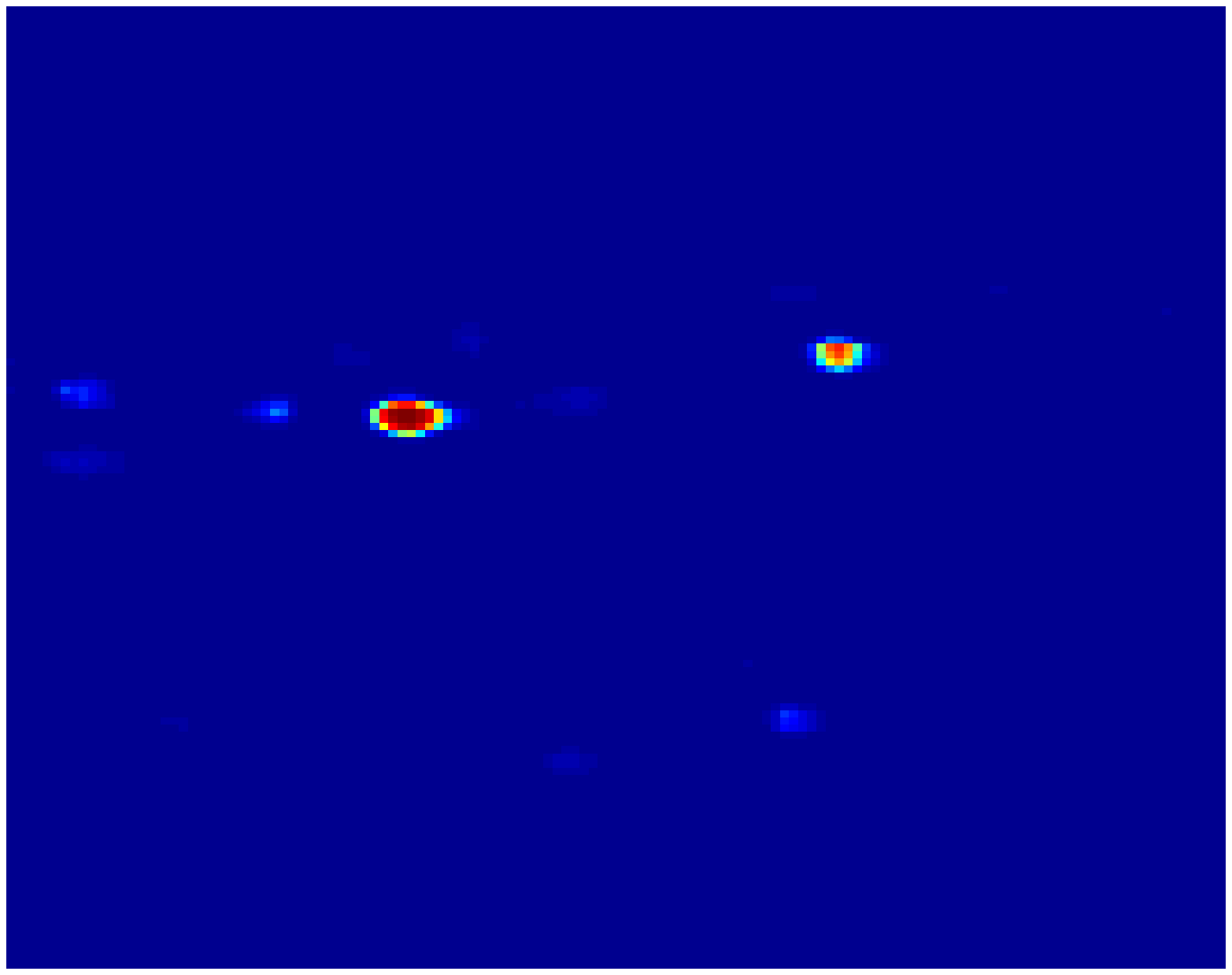}
\\
& `car' score, $\alpha = 0.33$ & `car' score, $\alpha = 0.57$\\
\includegraphics[width=\sz\columnwidth,height=\sz\columnwidth]{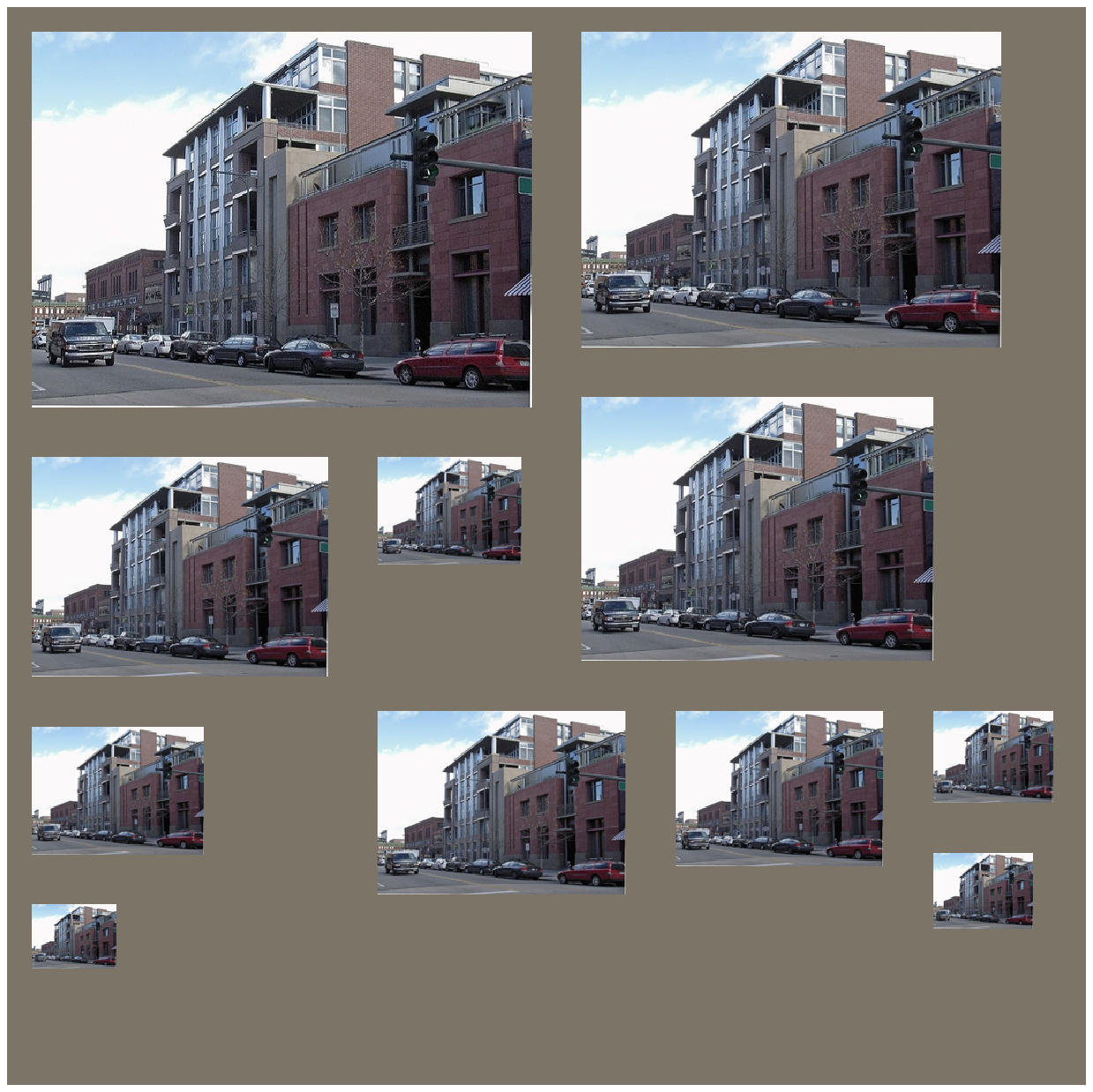}&
\includegraphics[width=\sz\columnwidth,height=\sz\columnwidth]{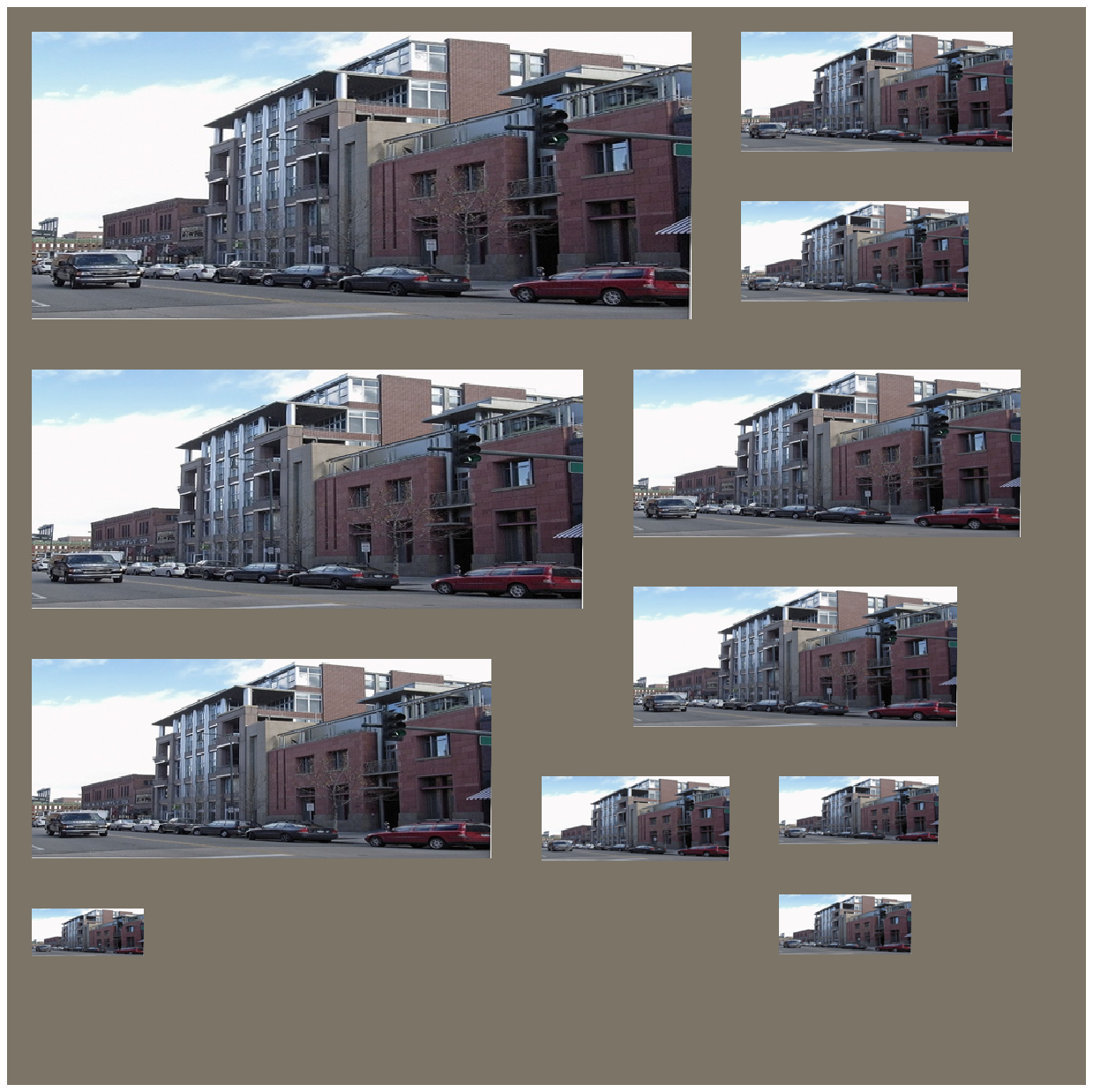}&
\includegraphics[width=\sz\columnwidth,height=\sz\columnwidth]{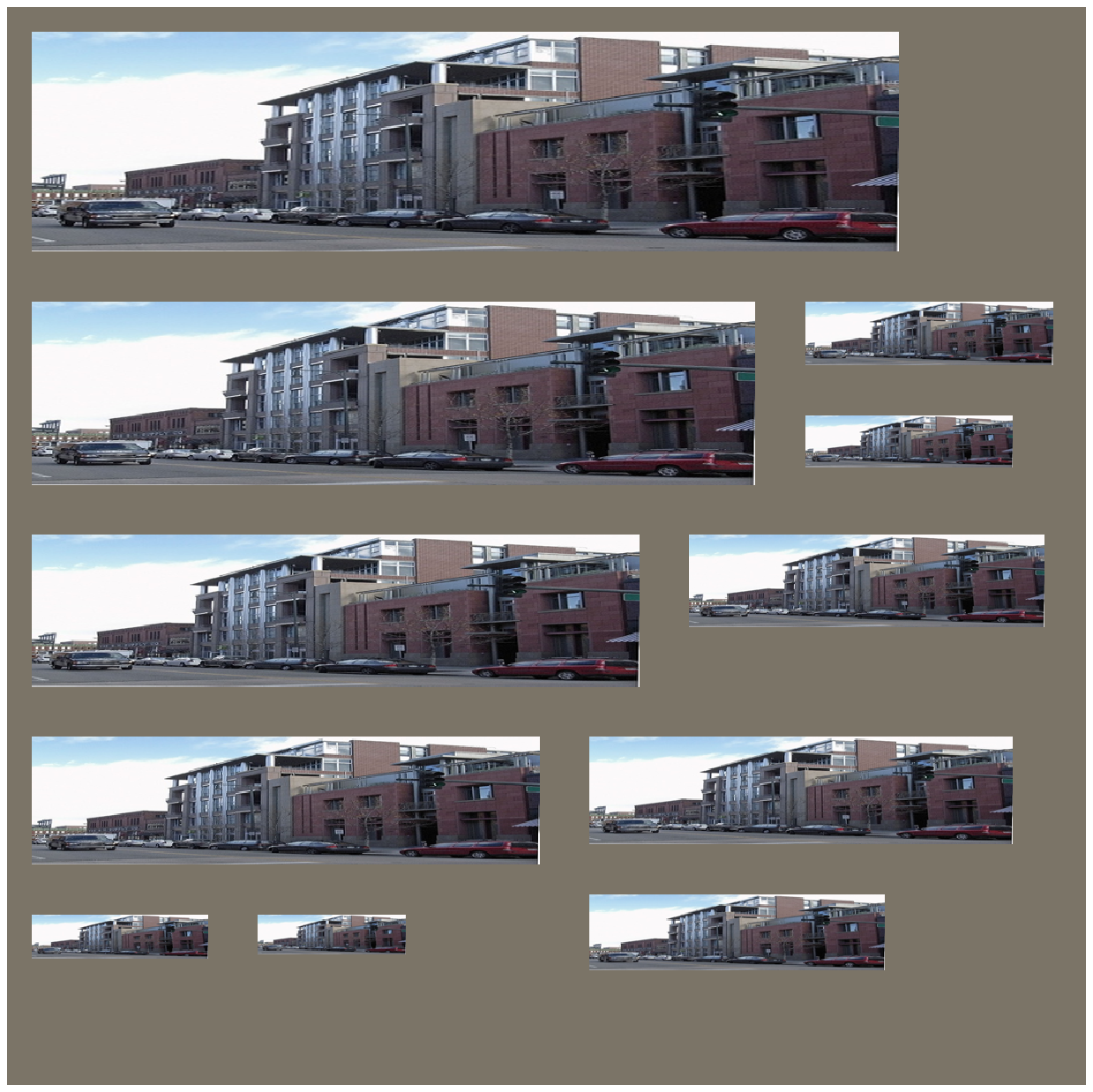}\\
Patchwork, $\alpha = 1$ & Patchwork, $\alpha = 1.73$ & Patchwork, $\alpha = 3.00$\\
\includegraphics[width=\sz\columnwidth,height=\sz\columnwidth]{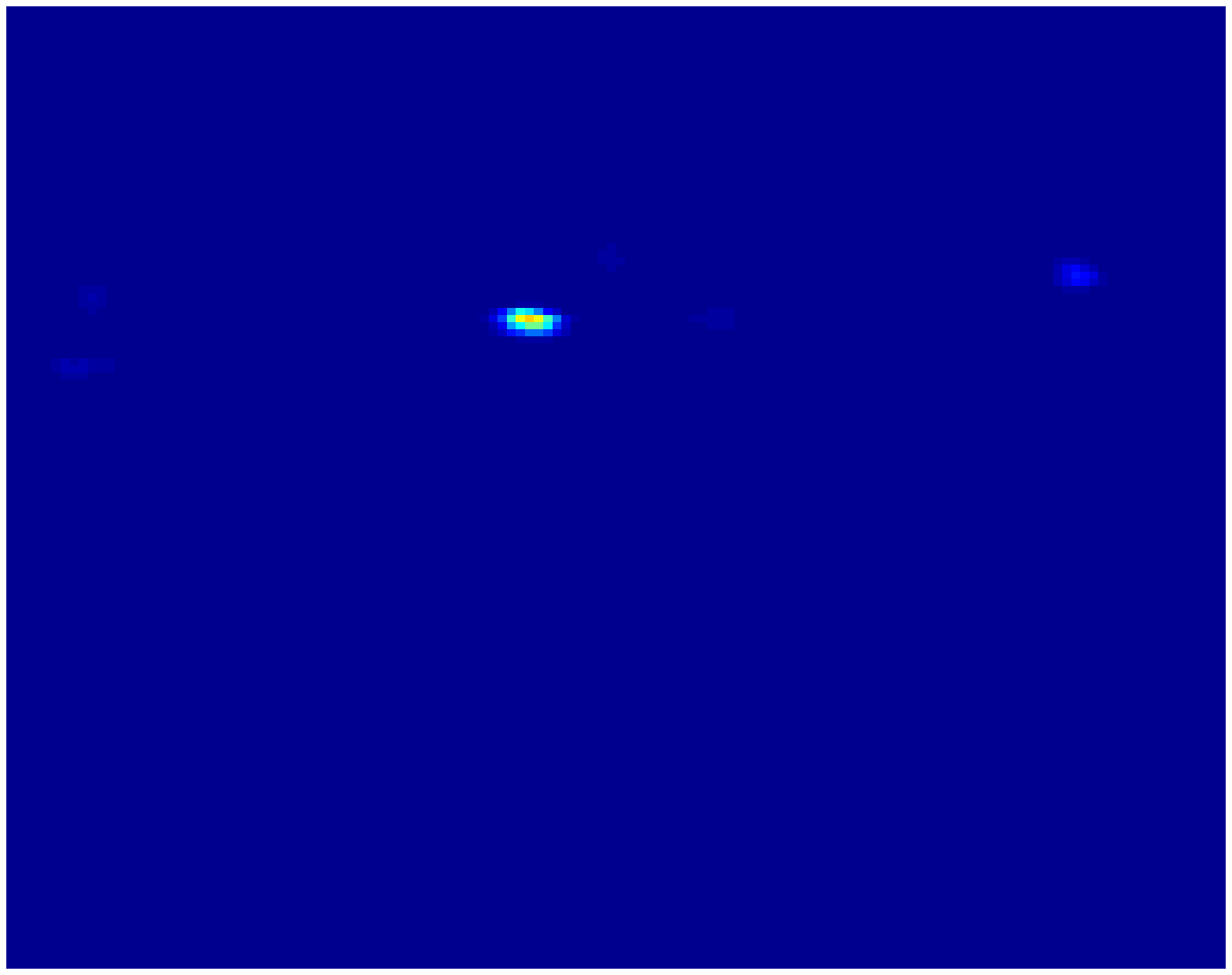}&
\includegraphics[width=\sz\columnwidth,height=\sz\columnwidth]{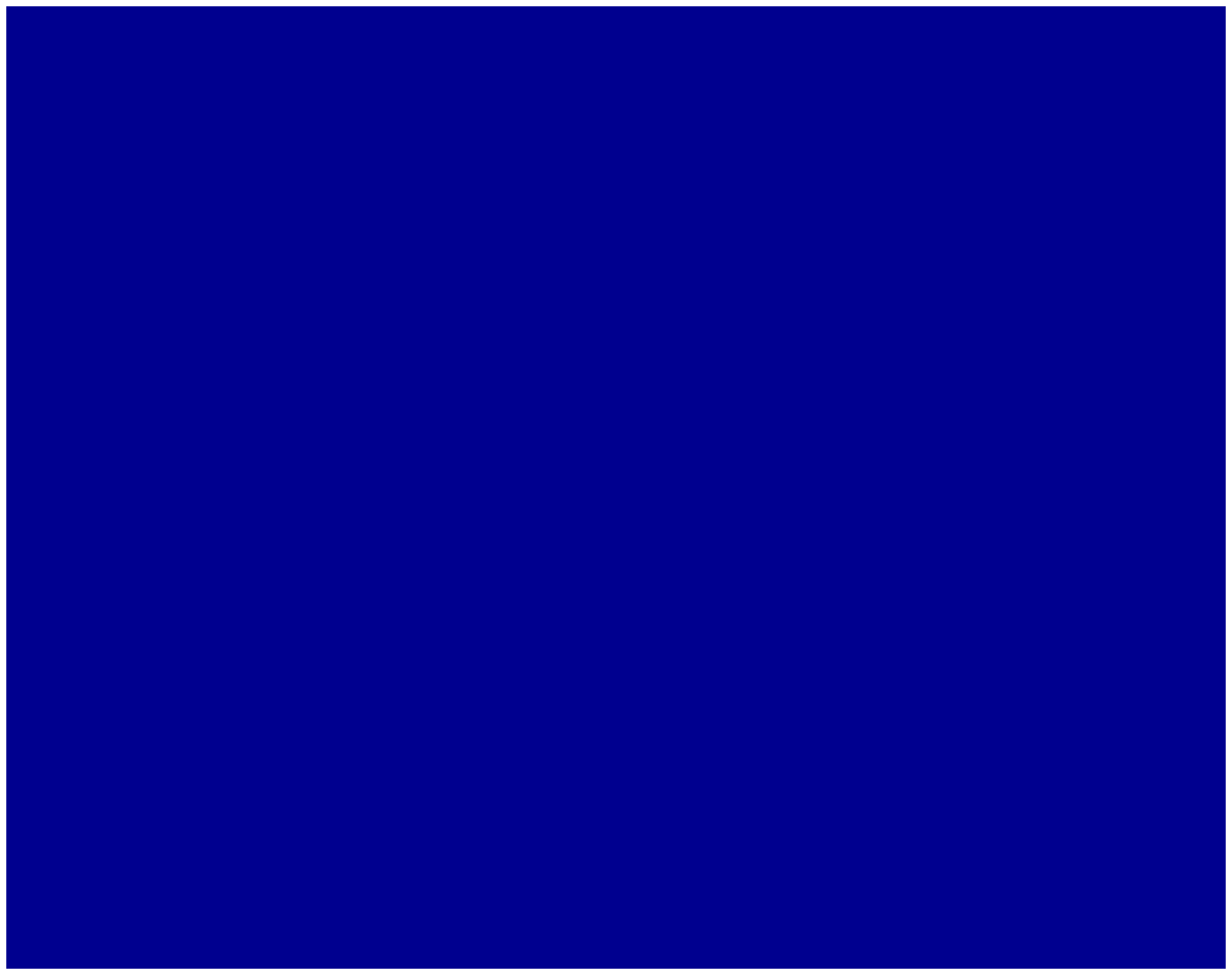}&
\includegraphics[width=\sz\columnwidth,height=\sz\columnwidth]{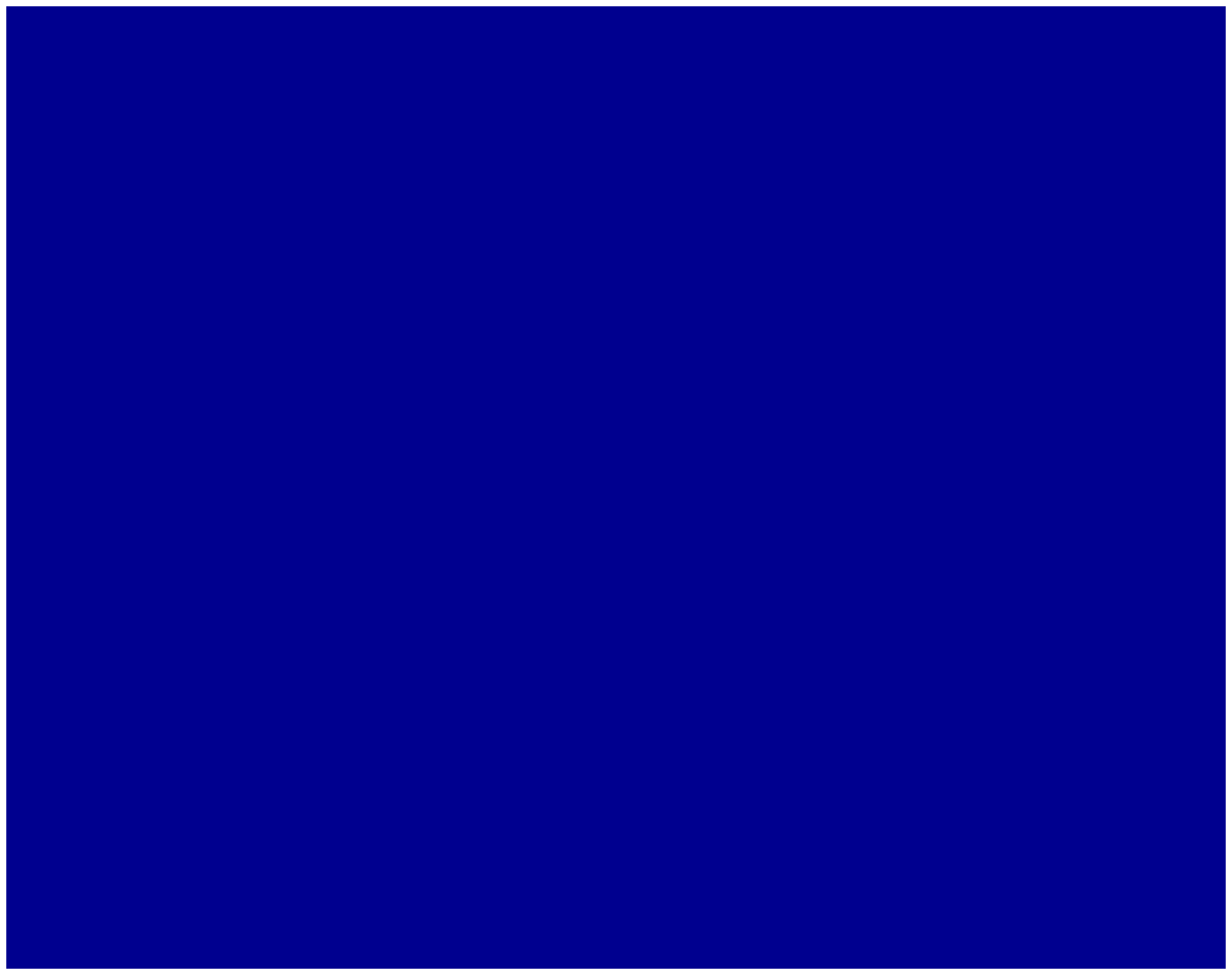}\\
 `car' score, $\alpha = 1.00$ & `car' score, $\alpha = 1.73$ &  `car' score, $\alpha = 3.00$ \\
\end{tabular}
\caption{Patchwork images (even rows) and car detector scores (odd rows) used for
  explicit position, scale, and aspect ratio (`$alpha$') search. We observe that the score is maximized at patchwork positions (i.e. image positions, scale, and aspect ratio combinations)  corresponding to  square-shaped cars. }
 \label{fig:pw}
\end{figure*}

\subsection{DCNN training for sliding window detection}
We deviate from the newtork fine-tuning used in the RCNN system \cite{GDDM14}
in two ways: firstly, we do not rely on the Selective Search \cite{ssearch}
region proposals to gather training samples; and, secondly, we modify the
network's structure to  process smaller images, and, resultantly,
include less parameters.  We detail these two modifications below.

\subsubsection{Model Training}
The RCNN system of \cite{GDDM14}  adapts a network trained for the Imagenet
classification task to object detection on Pascal VOC. For this, the authors
use the regions proposed by selective search to generate positive and negative
training samples, for each of the 20 categories of Pascal VOC; if a region has
an Intersection-over-Union  (IoU) above $.5$ for any bounding box of a class
it is declared as being a positive example for that class, or else it is a
negative example. These examples are used in  a network `fine-tuning' stage,
which amounts to running back-propagation with these training samples.

Rather than relying on Selective Search to provide training samples, we
exploit the dense sampling of positions, scales, and aspect ratios of our
algorithm. This allows us to use substantially cleaner examples, and train
with a higher IoU threshold for positives.

\begin{figure*}
\centering
\begin{tabular}{c c c c}
\includegraphics[width=.5\columnwidth]{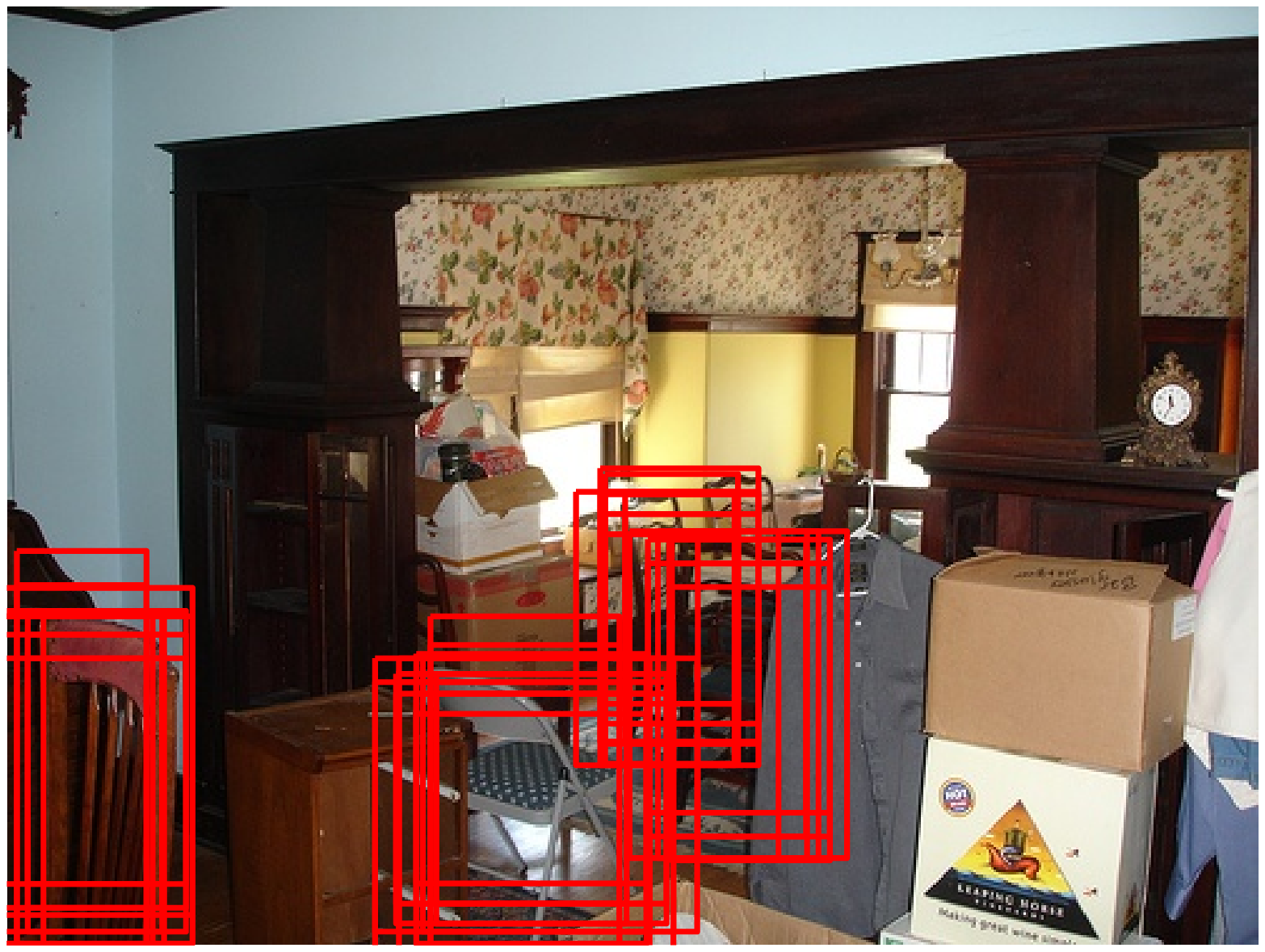}&
\includegraphics[width=.5\columnwidth]{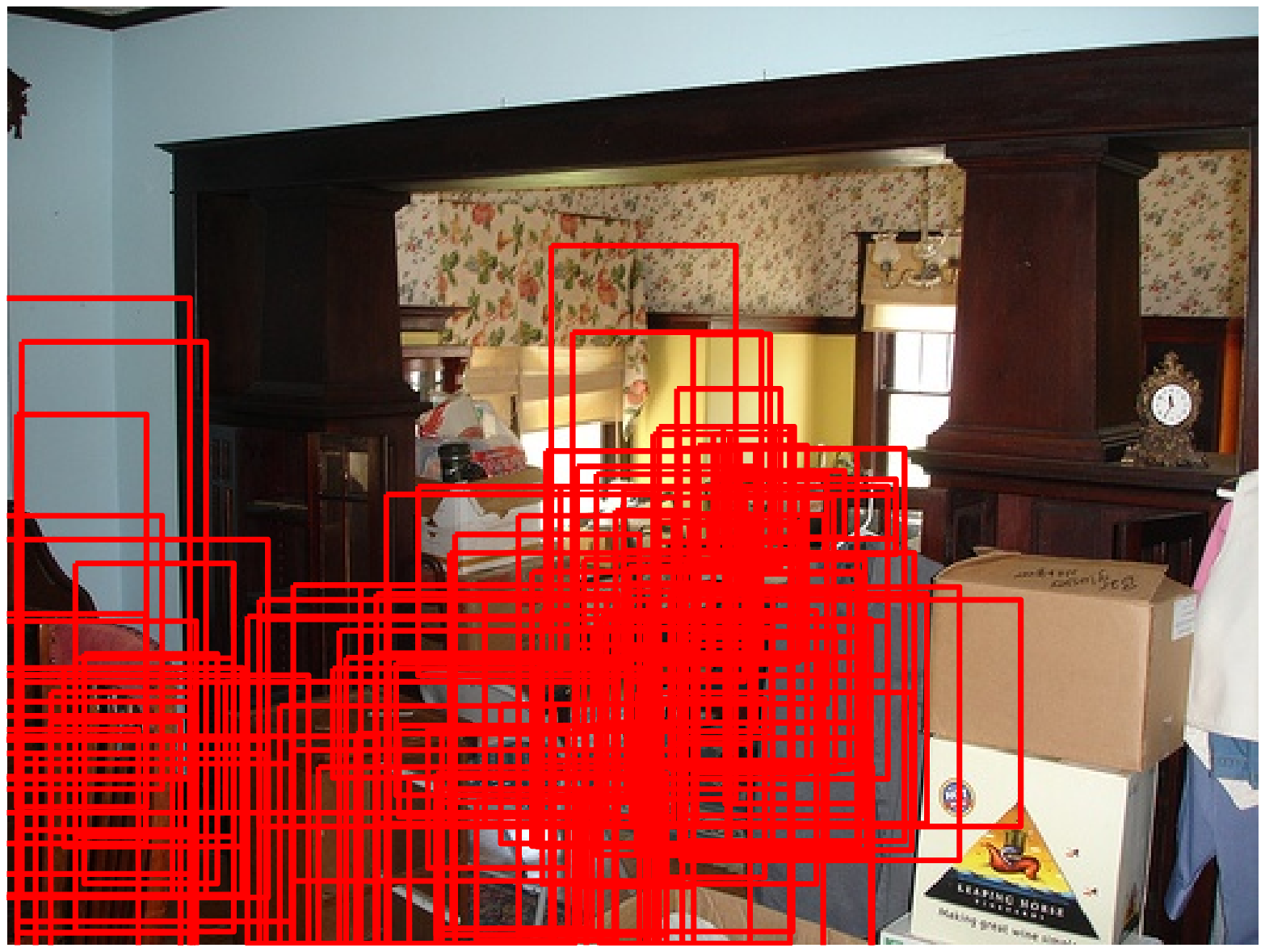} &
\includegraphics[width=.5\columnwidth]{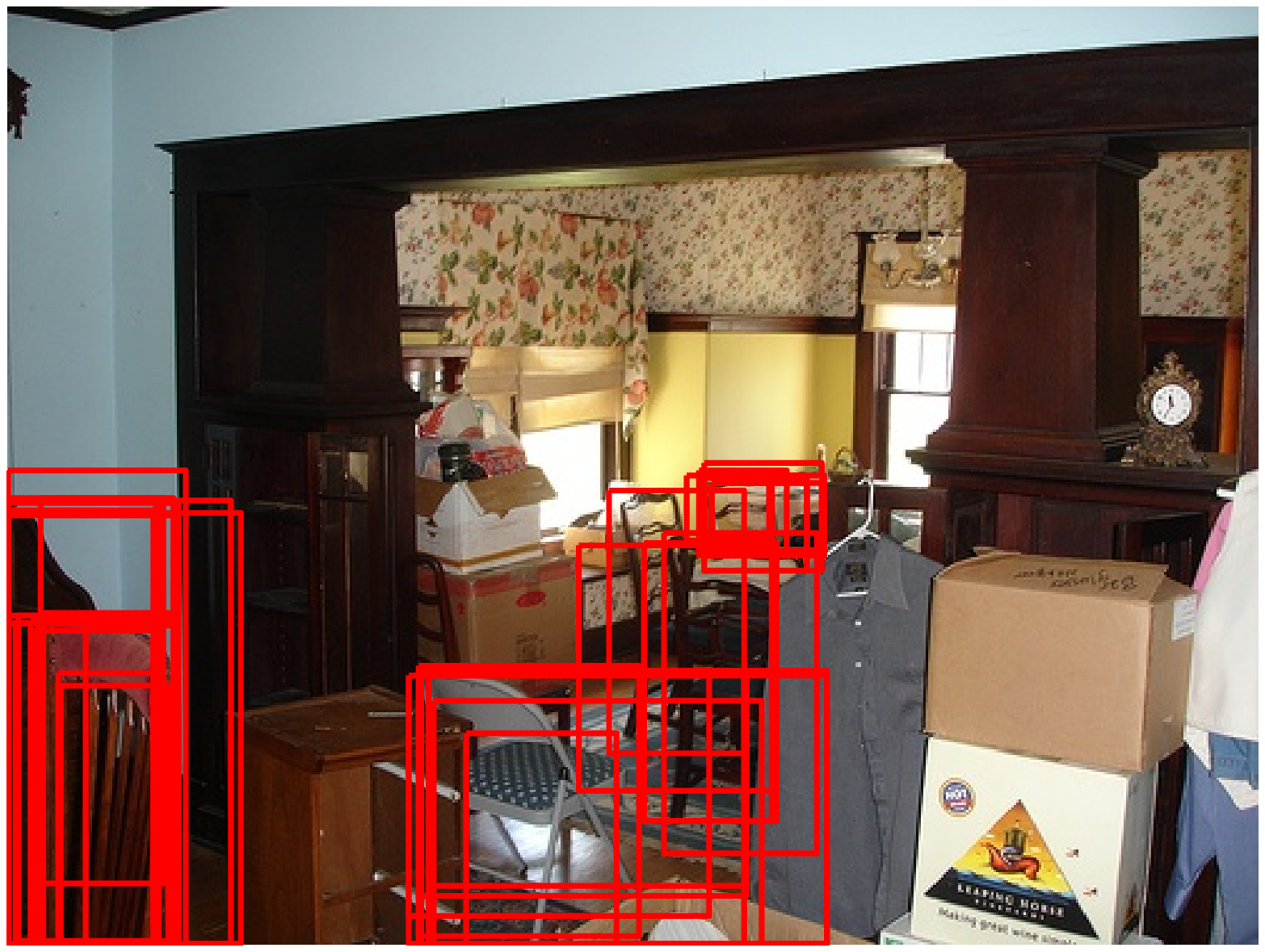}&
\includegraphics[width=.5\columnwidth]{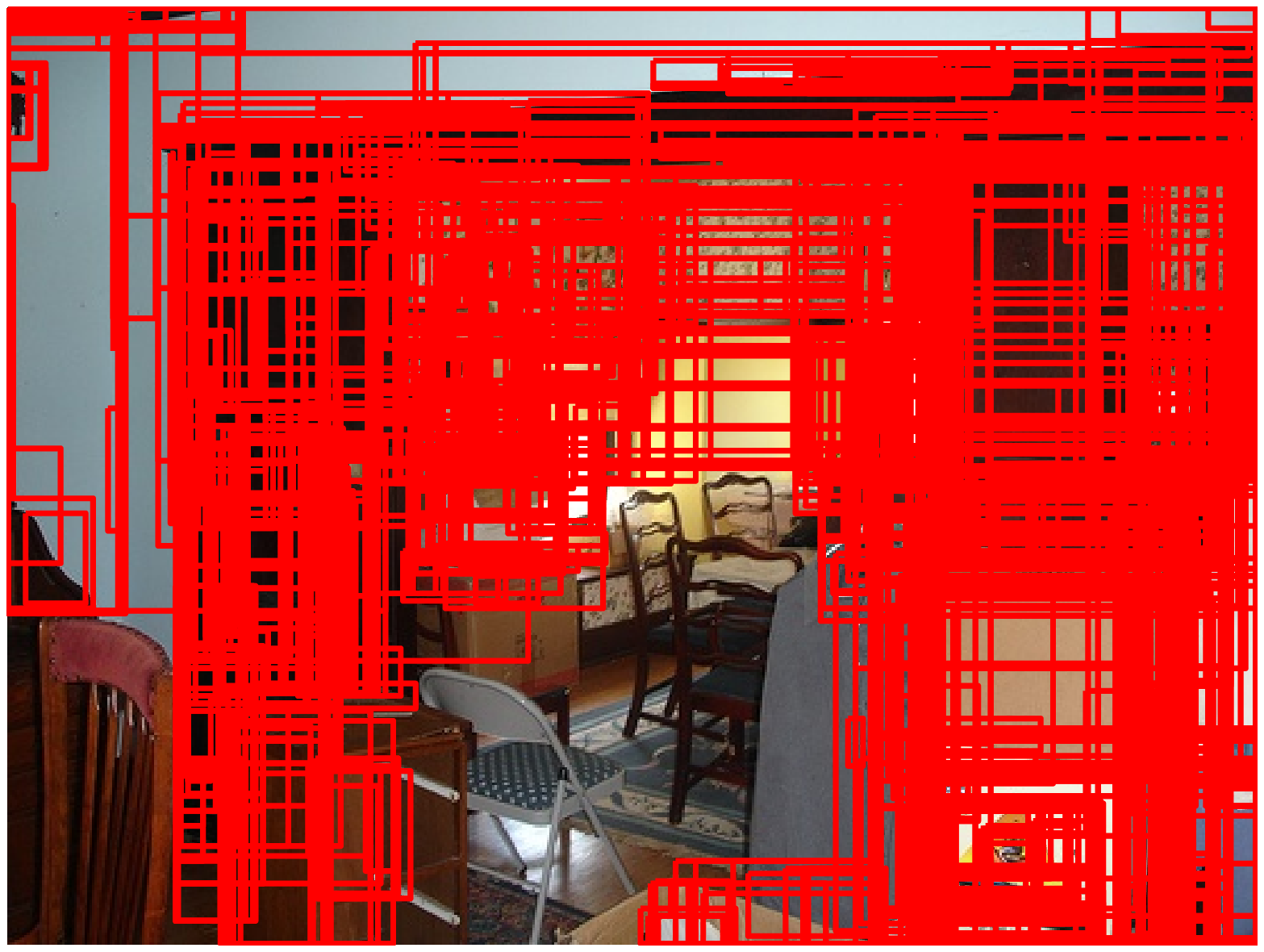}\\
a&
b &
c&
d\\
\multicolumn{2}{c}{
Sliding window positives (left)
and  negatives (right)} & 
\multicolumn{2}{c}{
Selective search positives (left)
and  negatives (right).}
\end{tabular}
\caption{Bounding boxes used for network finetuning. Our sliding window detector can use many more bounding boxes as positives (a) and negatives (b). The training samples available from selective search (c) are fewer, forcing the training to use poorly localized positives (d). When finetuning our network only with the latter examples performance would deteriorate.}
\label{fig:posneg}
\end{figure*}

In particular, as illustrated in Figure~\ref{fig:posneg}, we keep track of all
the windows that would be visited by our sliding window detector; given a
ground-truth bounding box, we randomly pick 30 of those that have an IoU score
above $0.7$ with it; if we have less than 30, we decrease the threshold to
$0.6$ and if we still cannot find as many we set the threshold finally to
$0.5$.  Similarily, for every positive bounding box we sample 200 negative
boxes that have an IoU score between $0.2$ and $0.5$ - aiming at `teaching'
our classifier what a poor localization looks like.

We have verified that doing this, rather than using selective search windows
gives us  clearly better detector scores, both visually and quantitatively. We
consider this to be one of the advantages of using a sliding window detector,
namely that we do not need to rely on an external region proposal module for
training.

\subsubsection{Re-purposing Classification Networks for Image Detection}

Herein we describe how we have re-purposed the publicly available state-of-art
16 layer classification network of \cite{SiZi14} (VGG-16) into an efficient
and effective component of our sliding window detector.

\paragraph{Dense sliding window feature extraction with the hole algorithm}

Dense spatial score evaluation is instrumental in the success of our CNN
sliding window detector. 

As a first step to implement this, we convert the fully-connected layers
of VGG-16 into convolutional ones and run the network in a convolutional
fashion on the patchwork. However this is not enough as it yields very
sparsely computed detection scores (with a stride of 32 pixels). To compute
scores more densely at our target stride of 8 pixels, we develop a variation
of the method previously employed by \cite{GCMG+13, SEZM+14}. We skip
subsampling after the last two max-pooling layers in the network of
\cite{SiZi14} and modify the convolutional filters in the layers that follow
them by introducing zeros to increase their length (\by{2}{} in the last three
convolutional layers and \by{4}{} in the first fully connected layer). We can
implement this more efficiently by keeping the filters intact and instead
sparsely sample the feature maps on which they are applied on using a stride
of 2 or 4 pixels, respectively. This approach is known as the `hole algorithm'
(`atrous algorithm') and has been developed before for efficient computation
of the undecimated wavelet transform \cite{Mall99}. We have implemented this
within the Caffe framework by adding to the \textsl{im2col} function (it
converts multi-channel feature maps to vectorized patches) the option to
sparsely sample the underlying feature map.

\paragraph{Shrinking the receptive field of neural networks}

Most recent DCNN-based image recognition methods rely on networks pre-trained
on the Imagenet large-scale classification task. These networks typically have
large receptive field size, \by{224}{224} in the case of the VGG-16 net we
consider. We have found this receptive field size to be too large to allow
good localization accuracy (unless one uses heavily zoomed-in versions of the
image). Moreover, after converting the network to a fully convolutional one,
the first fully connected layer has 4,096 filters of large \by{7}{7} spatial
size and becomes the computational bottleneck in our sliding window detector
system.

We have addressed both of these serious practical problems by spatially
subsampling the first FC layer to \by{4}{4} spatial size. This has reduced the
receptive field of the network down to \by{128}{128} pixels and has reduced
computation time for the first FC layer by 3 times.

\section{Conclusions}

This paper examines multiple facets of invariance in the context of deep
convolutional networks for visual recognition. First, we have proposed a new
epitomic convolutional layer which acts as a substitute to a pair of
consecutive convolution and max-pooling layers, and shown that it brings
performance improvements and exhibits better behavior during training. Second,
we have demonstrated that treating scale and position as latent variables and
optimizing over them during both training and testing yields significant
image classification performance gains. Pushing scale and position search
further, we have shown promising  results which
suggest that DCNNs can be efficient and effective for dense sliding window
based object detection. Further pursuing this topic is the main direction of
our future work.

\paragraph{Reproducibility} We implemented the proposed methods by extending
the excellent Caffe software framework \cite{Jia13}. When this work gets
published we will publicly share our source code and configuration files with
exact parameters fully reproducing the results reported in this paper.

\paragraph{Acknowledgments} We gratefully acknowledge the support of NVIDIA
Corporation with the donation of GPUs used for this research.


\bibliographystyle{ieee}
\bibliography{IEEEabrv,biblio_preamble_abrv,biblio_gpapan,nvidia}


\end{document}